\pdfoutput=1

\documentclass[11pt]{article}

\usepackage[]{acl}

\usepackage{times}
\usepackage{latexsym}
\usepackage{multirow}
\usepackage{booktabs}
\usepackage{graphicx}

\usepackage{amsmath}
\usepackage{amssymb}
\usepackage{adjustbox}     

\usepackage{microtype}
\usepackage{graphicx}
\usepackage{amsmath}
\usepackage{lipsum}
\usepackage{algorithm}
\usepackage{algpseudocode}
\usepackage{amsfonts}
\usepackage{subcaption}
\usepackage{comment}
\usepackage{url}
\usepackage{multirow}
\usepackage{multirow}
\usepackage{booktabs}
\DeclareUnicodeCharacter{2212}{-}
\usepackage{makecell}

\usepackage{bm}

\usepackage[T1]{fontenc}

\usepackage[utf8]{inputenc}

\usepackage{microtype}

\title{A Holistic Framework for Analyzing the COVID-19 Vaccine Debate}

\author{
   Maria Leonor Pacheco$^{1*}$ \,\,
   Tunazzina Islam$^{1*}$ \,\,
   Monal Mahajan$^1$   \,\, \\
   \textbf{Andrey Shor}$^1$  \,\,
   \textbf{Ming Yin}$^1$ \,\,
   \textbf{Lyle Ungar}$^2$ \,\,
   \textbf{Dan Goldwasser}$^1$ \\
    $^1$CS, Purdue University, USA,~~
    $^2$CIS, University of Pennsylvania, USA \\
    \texttt{\{pachecog,islam32,mmahaja,ashor,mingyin,dgoldwas\}@purdue.edu} \\
    \texttt{ungar@cis.upenn.edu}
}

\begin{document}

\maketitle
\begin{abstract}
The Covid-19 pandemic has led to infodemic of low quality information leading to poor health decisions. Combating the outcomes of this infodemic is not only a question of identifying false claims, but also reasoning about the decisions individuals make.
In this work we propose a holistic analysis framework connecting stance and reason analysis, and fine-grained entity level moral sentiment analysis. We study how to model the dependencies between the different level of analysis and incorporate human insights into the learning process. Experiments show that our framework provides reliable predictions even in the low-supervision settings.
\end{abstract}

\footnotetext[1]{Equal contribution}

\section{Introduction}






One of the unfortunate side-effects of the Covid-19 pandemic is a global infodemic flooding social media with low quality and polarizing information about the pandemic, influencing public perception on it~\cite{tagliabue2020pandemic}. As studies have shown~\cite{montagni2021acceptance}, these influences have clear real-world implications, in terms of public acceptance of treatment options, vaccination and prevention measures. 

Most computational approaches tackling the Covid-19 infodemic view it a misinformation detection problem. In other words, they look at identifying false claims and analyzing reactions to them on social media ~\cite{hossain-etal-2020-covidlies,alam2021fighting,weinzierl2021misinformation}. This approach, while definitely a necessary component in fighting the infodemic, does not provide policy makers and health-professionals with much needed information, characterizing the reasons and attitudes that underlie the health and well-being choices individuals make. 

\begin{figure}[t]
    \centering
    \includegraphics[width=\columnwidth]{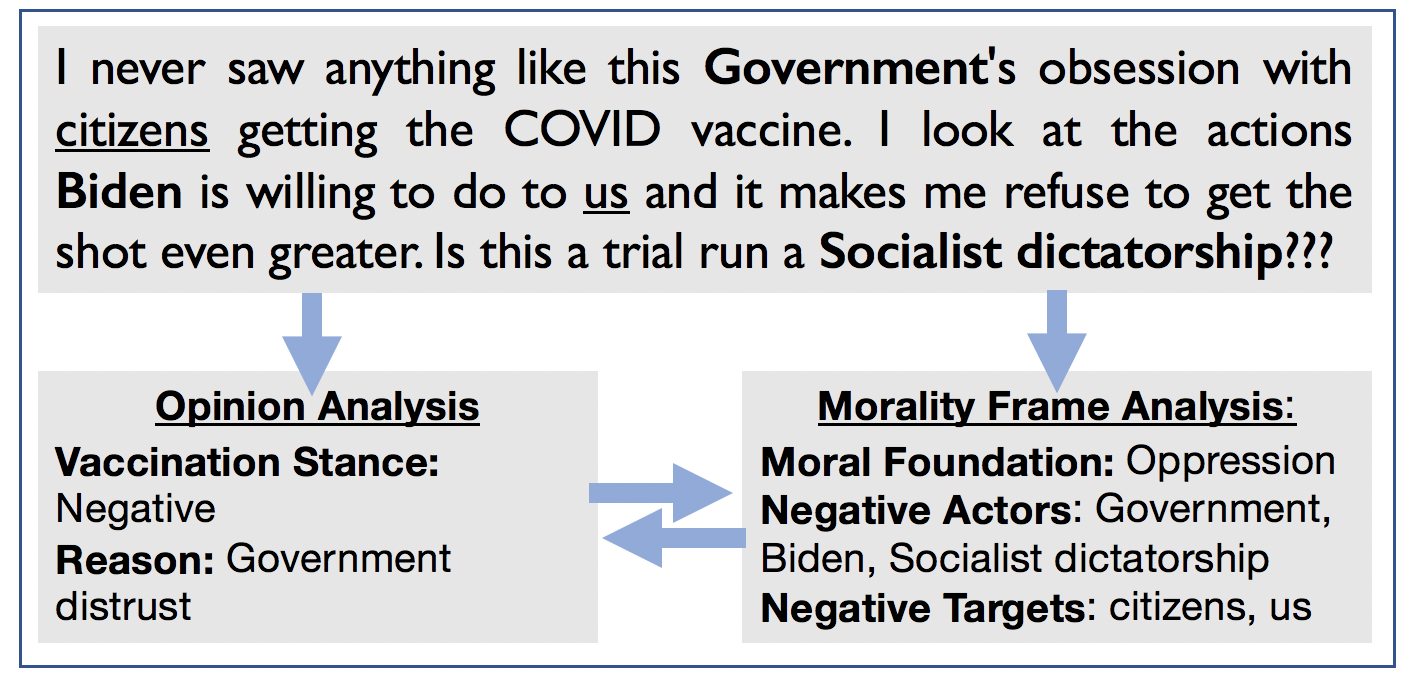}
    \caption{Holistic Analysis Framework of Social Media Posts, Connecting entity-level Moral Perspectives, Stance and Arguments Justifying it.}
    \label{fig:exampleAnalysis}
    \vspace{-10pt}
\end{figure}

Our goal in this paper is to suggest a holistic analysis framework, providing multiple interconnected views of the opinions expressed in text. We specifically focus on a timely topic, attitudes explaining vaccination hesitancy. Figure~\ref{fig:exampleAnalysis} describes an example of our framework. Our analysis identifies the  \textit{stance} expressed in the post ({\small\texttt{anti-vaccination}}) and the \textit{reason} for it ({\small\texttt{distrust of government}}). Given the ideologically polarized climate of social media discussion on this topic, we also aim to characterize the moral attitudes expressed in the text ({\small\texttt{oppression}}), and how different entities mentioned in it are perceived ({\small\texttt{``Biden, Government'' are oppressing, ``citizens, us'' are oppressed}}). When constructing this framework we tackled three key challenges.  

\textbf{1. How should these analysis dimensions be operationalized?} While stance prediction is an established NLP task, constructing the space of possible arguments justifying stances on a given topic, and their identification in text, are still  open challenges. \textit{We take a human-in-the-loop approach to both problems}. We begin by defining a seed set of relevant arguments based on data-driven studies~\cite{weinzierl2021misinformation,sowa2021covid}, where each reason is defined by a single exemplar sentence. In a sequence of interactions, we use a pre-trained textual inference model to identify paraphrases in a large collection of Covid-19 vaccination tweets, and present a visualization of the results to humans. Humans then perform an error analysis, and either add more sentences to help characterize the existing reason better, or add and characterize additional reasons, based on examples retrieved from the large corpus. We explain this process in detail in Section~\ref{sec:opinions}.

Our morality analysis is motivated by social science studies~\cite{pagliaro2021trust,diaz2021reactance,chan2021moral} that demonstrate the connection between moral foundation preferences~\cite{haidt2007morality,graham2009}
 and Covid-related health choices. For example, studies show that the endorsement of the \textit{fairness} and \textit{care} moral foundations is correlated with trust in science. To account for fine-grained patterns, we adapt the recently proposed morality-frame formalism~\cite{roy-etal-2021-identifying} that identifies moral roles associated with moral foundation expressions in text. These roles correspond to actor/target roles (similar to agent/patient) and positive or negative polarity, which should be understood in the context of a specific moral foundation. In Figure~\ref{fig:exampleAnalysis} \textit{``Biden''} is the negative actor in the context of Oppression, making him the oppressor. We explain this formalism in Section~\ref{sec:morality}.
  
\textbf{2. How should the dependencies between these dimensions be captured and utilized?} The combination of stance, reason and moral attitudes provides a powerful source of information, allowing us to capture the moral attitudes expressed in the context of different stances and their reasons.  These connections can also be used to help build expectations about likely attitudes in the context of each stance. As a motivating example, consider the reason {\small\texttt{``distrust in government''}}, which can be associated with the  {\small\texttt{``oppression''}} moral foundation \textit{only} when its actor is an entity related to government functions (rather than oppression from Covid-19 illness). We model these expectation as a probabilistic inference process~\cite{pacheco-goldwasser-2021-modeling}, by incorporating consistency constraints over the judgements made by our model, and predicting the most likely analysis jointly, consisting of all analysis dimensions. The full model, described using a declarative modeling language, is provided in Section~\ref{sec:model}.
 
\textbf{3. How can text analysis models be adapted to this highly dynamic domain, without extensive and costly manual annotation?} While our analysis in this paper focuses on a specific issue, vaccination hesitancy, we believe that our analysis framework should be easily adaptable to new issues. Relying on human insight to characterize and operationalize stance and reason identification is one aspect, that characterizes \textit{issue-specific} considerations. Moral Foundation Theory, by its definition, abstracts over specific debate topics, and offers a general account for human morality. However, from a practical perspective, models for predicting these highly abstract concepts are trained on data specific to a debate topic and might not generalize well. Instead of retraining the model from scratch, we hypothesize that given an initial model constructed using out-of-domain data, and a small amount of in-domain labeled data, we can obtain acceptable performance by modeling the interaction between reasons, stances and moral foundations. We study these settings, along with the fully supervised setting in Section~\ref{sec:experiments}. 

The data, code and tools used in this paper are publicly available\footnote{\url{https://gitlab.com/mlpacheco/covid-moral-foundations}}. 


\section{Opinion Analysis}\label{sec:opinions}

To analyze opinions about the Covid-19 vaccine, we model the vaccination stance expressed in each tweet (i.e. pro-vaccine, anti-vaccine, neutral) and the underlying reason behind such stance. For example, in Figure~\ref{fig:exampleAnalysis} the tweet expresses an anti-vaccine stance, and mentions their distrust of the Biden administration as the reason to take this stance. 

There are three main challenges involved in this analysis: 1) predicting the stance, 2) constructing the space of possible reasons, and 3) mapping tweets to the relevant reasons. Stance prediction is an established NLP classification task~\cite{glandt-etal-2021-stance}. However, uncovering latent themes from text automatically remains an open challenge, traditionally approached using noisy unsupervised techniques such as topic models \cite{zamani-etal-2020-understanding}, or by manually identifying and annotating them in text~\cite{hasan-ng-2014-taking}.


Instead, we combine computational and qualitative techniques to uncover the most frequent reasons cited for pro and anti vaccination stances. We build on previous health informatics studies that characterized the arguments made against the Covid-19 vaccine in social media \cite{wawrzuta2021arguments}. In this work, researchers come up with a code-book of 12 main themes, frequently used as reasons to refuse or cast doubt on the vaccine. 
\textbf{We propose an interactive, humans-in-the-loop protocol} to learn representations for these 12 initial reasons, ground them in data, evaluate their quality, and refine them to better capture the discussion. 
%
%
%
%
%
%
To do this, we build a tool to explore repeating arguments and their reasons in the Covid-19 vaccine debate. The tool consists of an interactive Google Colab notebook equipped with a custom API to query current arguments, ground them in data, and visualize them. To initialize the system, we use the 12 reasons suggested by \citet{wawrzuta2021arguments}, and represent them using the one-sentence explanation provided. Our main goal is to ground these reasons in a set of approximately 85,000 unlabeled tweets about the Covid-19 vaccine (details in Section \ref{sec:data}). To map tweets to reasons, we use the similarity between their SBERT embeddings \cite{reimers-gurevych-2019-sentence}. The interaction is centered around the operations outlined in Table \ref{tab:operations}. Intuitively, the first six operations allow humans to diagnose how reasons map to text, and the last three allows them to act on the result of this diagnosis, by adding and removing reasons, and modifying the phrases characterizing each reason.

\begin{table}[t]
\begin{center}
 \scalebox{0.7}{\begin{tabular}{>{\arraybackslash}m{10cm}}
 \toprule
\texttt{\textbf{show\_reasons()}} lists the current list of reasons (e.g. Government Distrust, Natural Immunity.)  \\
\hline
\texttt{\textbf{show\_closest\_tweets(reason, K)}} lists the \texttt{K} tweets closest to a given \texttt{reason}, based on their embedding similarity. \\
\hline
\texttt{\textbf{wordcloud(reason)}} Renders a word cloud to visualize the arguments associated to a given reason, based on bigram and trigram TF-IDF features.  \\
\hline
\texttt{\textbf{show\_assignments(threshold)}} Renders a bar plot showing the assignment of tweets to reasons, based on embedding similarity. An optional threshold can be used to limit assignments.\\
\hline
 \texttt{\textbf{tsne(threshold)}} Renders a visualization of the reason clusters in a 2D map. Threshold is optional. \\
\hline 
\texttt{\textbf{silhouette\_score(threshold)}} Measures the overlapping degree between clusters. Threshold is optional.  \\
\hline 
\texttt{\textbf{add\_reason(reason, phrase)}} Adds a new reason with a phrase that characterizes it in natural language \\
\hline
\texttt{\textbf{remove\_reason(reason)}} Removes a given reason \\
\hline
\texttt{\textbf{add\_phrase(reason, phrase)}} Adds an additional phrase to an existing reason. \\
\bottomrule
\end{tabular}}
\caption{Interactive API Operations}
\label{tab:operations}
\end{center}
\vspace{-10mm}
\end{table}

We follow a simple protocol during interaction, where three human coders use the operations above to explore the initial reasons. The coders start by looking at the global picture: the reasons distribution, the 2D visualizations~\cite{vanDerMaaten2008} and the silhouette score~\cite{Rousseeuw87silhouetteCluster}. Then, they query the reasons one by one, looking at the word cloud (characterizing the distribution of short phrases over all texts assigned to the reason) and the 10 closest tweets to each reason. Following these observations, there is a discussion phase in which the coders follow a thematic analysis approach~\cite{thematic_analysis} to uncover the overarching themes that are not covered by the current set of reasons, as well as the argumentation patterns that the method fails to identify. Then, they are allowed to add and remove reasons, as well as explanatory phrases for them in natural language. Every time a reason or phrase is added or removed, all tweets are reassigned to their closest reasons. This process was done over two one-hour sessions. The coders were NLP and Computational Social Science researchers, two female and one male, between the ages of 25 and 40. 

In the first session, the coders focused on adding new reasons and removing reasons that were not prevalent in the data. For example, they noticed that the initial set of reasons contained mostly anti-vaccine arguments, and added a positive reason for each negative reason (e.g. \textit{government distrust} $\Rightarrow$ \textit{government trust}). In addition to this, they broke down the reason "Conspiracy Theory" into specific conspiracy theories, such as \textit{Bill Gates' micro chip}, \textit{the vaccine contains fetal tissue}, and \textit{the vaccine makes you sterile}.
They also removed infrequent reasons, such as \textit{the swine flu vaccine}. 
The final set of reasons can be observed in Table \ref{tab:reasons}.

\begin{table}[t]
\begin{center}
 \scalebox{0.7}{\begin{tabular}{>{\arraybackslash}m{1cm}|>{\arraybackslash}m{9cm}}
 \toprule
 \textsc{\textbf{Pro Vax}} & government distrust, vaccine dangerous, covid fake, vaccine oppression, pharma bad, natural immunity effective, vaccine against religion, vaccine does not work, vaccine not tested, bill gates' micro chip, vaccine tested on dogs, vaccine has fetal tissue, vaccine makes you sterile \\
 \hline
 \textsc{\textbf{Anti Vax}} & government trust, vaccine safe, covid real, vaccine not oppression, pharma good, natural immunity ineffective, vaccine not against religion, vaccine works, vaccine tested \\
 \bottomrule
\end{tabular}}
\caption{Resulting Reasons}
\label{tab:reasons}
\end{center}
\end{table}

In the second session, the coders focused on identifying the argumentative patterns that were not being captured by the original reason explanations, and came up with overarching patterns to create new examples to improve the representation of the reasons. For example, in the case of the \textit{government distrust} reason, the coders found that phrases with strong words were needed (e.g. \textit{F the government}), examples that suggested that the government was "good at being bad" (e.g. \textit{the government strong record of screwing things up}), and examples with explicit negations (e.g. \textit{the government does not work logically}). Once patterns were identified, each coder contributed a set of 2 to 5 examples, which were introduced to the reason representation. 

In Appendix \ref{app:themes_phrases}, we include screenshots of the interactive notebook, and tables enumerating the full list derived patterns and phrases. To visualize the impact of interaction, we also show the overall distribution of reasons before and after interaction, and word clouds for a select set of reasons. The methodology and tool we developed are broadly applicable for diagnosing NLP models.
\section{Morality Frame Analysis}\label{sec:morality}

Moral Foundations Theory \cite{haidt2007morality} suggests that there are at least six basic foundations that account for the similarities and recurrent themes in morality across cultures, each with a positive and negative polarity (See Table \ref{tab:moral_foundations}). 

\begin{table}[t]
\begin{center}
 \scalebox{0.7}{\begin{tabular}{>{\arraybackslash}m{10cm}}
 \toprule
\textsc{\textbf{Care/Harm:}} Underlies virtues of kindness, gentleness, and nurturance.  \\
\hline
\textsc{\textbf{Fairness/Cheating:}} Generates ideas of justice, rights, and autonomy. \\
\hline
\textsc{\textbf{Loyalty/Betrayal:}} Underlies virtues of patriotism and self-sacrifice for the group. It is active anytime people feel that it’s ``one for all, and all for one.'' \\
\hline
\textsc{\textbf{Authority/Subversion:}} Underlies virtues of leadership and followership, including deference to legitimate authority and respect for traditions. \\
\hline
\textsc{\textbf{Purity/Degradation:}} Underlies religious notions of striving to live in an elevated, less carnal, more noble way. It underlies the widespread idea that the body is a temple which can be desecrated by immoral activities and contaminants. \\
\hline 
\textsc{\textbf{Liberty/Oppression:}} The feelings of reactance and resentment people feel toward those who dominate them and restrict their liberty. \\
\bottomrule
\end{tabular}}
\caption{Moral Foundations \cite{haidt2007morality}}
\label{tab:moral_foundations}
\end{center}
\end{table}

To analyze moral perspectives in tweets, we build on the definition of morality frames proposed by \citet{roy-etal-2021-identifying}, where moral foundations are regarded as frame predicates, and associated with positive and negative entity roles.  

While \citet{roy-etal-2021-identifying} defined different roles types for each moral foundation (e.g. \textit{entity causing harm}, \textit{entity ensuring fairness}), we aggregate them into two general role types: \textbf{actor} and \textbf{target}, each with an associated polarity (positive, negative). An \textbf{actor} is a ``do-er'' whose actions or influence results in a positive or negative outcome for the \textbf{target} (the ``do-ee''). For each moral foundation in a given tweet, we identify the ``entity doing good/bad'' (positive/negative actor) and ``entity benefiting/suffering'' (positive/negative target). For example, the statement ``We are suffering from the pandemic'' expresses \textbf{harm} as the moral foundation, where ``pandemic'' is a \textbf{negative actor}, and ``we'' is a \textbf{negative target} (i.e. the entity suffering from the actor's actions). There can be zero, one or multiple actors and targets in a given tweet. Entities can correspond to specific individuals or groups (e.g., I, democrats, people of a given demographic), organizations (e.g., political parties, CDC, FDA, companies), legislation or other political actions (e.g., demonstrations, petitions), disease or natural disasters (e.g., Covid, global warming), scientific or technological innovations (e.g., the vaccine, social media, the Internet), among others.

We break down the task of predicting morality frames into four classification tasks. For each tweet, our goal is to predict whether it is making moral judgement or not, and identify its prominent moral foundation. For each entity mentioned in the tweet, we predict whether it is a target or a role, and whether it has positive or negative polarity.



\section{Data Collection and Annotation}\label{sec:data}

    \begin{figure*}[ht]
    \centering
\begin{subfigure}[t]{0.7\columnwidth}
  \centering
  \includegraphics[width=\textwidth]{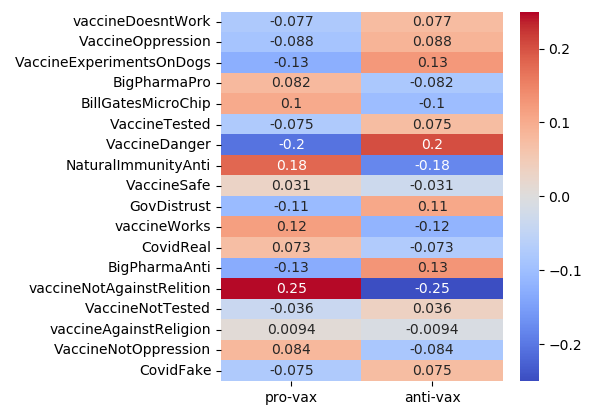}
  \caption{Reasons and Vax Stance}
  \label{fig:correl_reasons_stance}
\end{subfigure}
\begin{subfigure}[t]{0.8\columnwidth}
  \centering
 \includegraphics[width=\columnwidth]{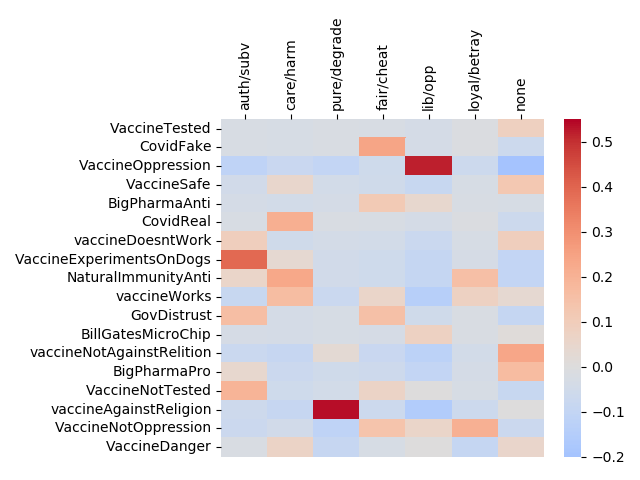}
    \caption{Reasons and Moral Foundations}
    \label{fig:correlations}
\end{subfigure}
\begin{subfigure}[t]{0.55\columnwidth}
  \centering
 \includegraphics[width=\columnwidth]{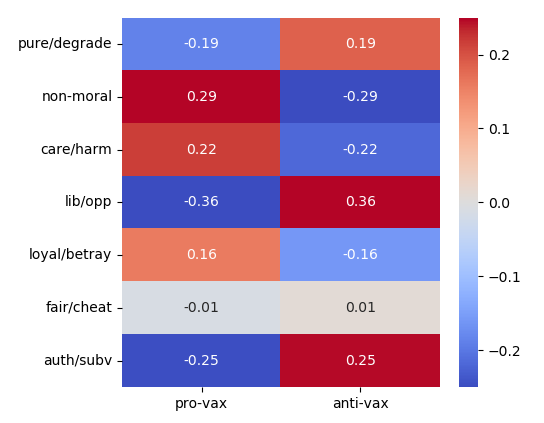}
    \caption{Moral Foundations and Vax Stance}
    \label{fig:correlations}
\end{subfigure}

\caption{Correlation Heatmaps}
\label{fig:correl_comparison}
\end{figure*}

There is no existing corpus of arguments about the Covid-19 vaccine annotated for morality frames and vaccination stance, so we collected and annotated our own. First, we searched for tweets between April
and October of 2021 mentioning specific keywords, such as \textit{covid vaccine} and \textit{vaccine mandate}. 
The full list of keywords, as well as the procedure to obtain them, can be seen in Appendix \ref{app:data_collection}.


Then, we created an exclusive web application for annotating our task. Moral foundation and vaccination stance labels can be annotated directly. To identify entities, annotators were able to highlight the relevant text spans, and choose its role label (i.e. positive/negative actor or target). 
We annotated our dataset using three in-house annotators pursuing a Ph.D. in Computer Science. We awarded the annotators \$ $0.75$ per tweet and bonus ($ 2 * \$ 0.75 = \$ 1.5 $) for completing two practice examples. Our work is IRB approved, and we follow their protocols.

To ensure quality work, we provided annotators with eight examples covering all six moral foundations and non-moral cases. Before starting the annotation task, the annotators had to read the instructions, go through the examples, and annotate two practice questions. The annotation interface, examples and practice questions can be seen in Appendix \ref{app:data_annotation_task}. 


\paragraph{Inter-annotator agreement} We calculated the agreement among annotators using Krippendorff’s $\alpha$ \cite{krippendorff2004measuring}, where $\alpha = 1$ suggests perfect agreement, and $\alpha = 0$ suggests chance-level agreement. We found $\alpha = 60.82$ for moral foundations, and $\alpha = 78.71$ for stance. For roles, we calculated the character by character agreement between annotations. For example, if one annotator marked ``Dr Fauci'' as a target in a tweet, and another
marked ``Fauci'', it was considered to be an agreement on the characters ``Fauci'' but
disagreement on ``Dr''. Doing this, we found $\alpha = 83.46$. When removing characters marked by all three annotators as "non-role", the agreement dropped to $\alpha = 67.15$. 

\paragraph{Resulting annotated dataset} We used a majority vote to get moral foundation and vaccination stance labels, and obtained 750 annotated tweets. Similarly, we defined a text span to be an entity mention E, having a moral role R and polarity P, in a tweet T, if it was annotated as such by at least two annotators. Our resulting dataset contains 891 (T,E,R,P) tuples. Statistics can be seen in Table \ref{tab:dataset}. 

\begin{table}[H]
    \centering
    \resizebox{\columnwidth}{!}{%
    \begin{tabular}{ll|llll}
    \toprule
        \textbf{\textsc{Moral}} &  \textbf{\textsc{Num.}} & \multicolumn{4}{c}{\textbf{\textsc{Vaccination Stance}}}\\
         \textbf{\textsc{Foundation}} & \textbf{\textsc{Tw.}} & \textsc{Pro} & \textsc{Anti} & \textsc{Neut} & \textsc{No Agree} \\
    \midrule
    Care/Harm & 96 & \textbf{77} & 17 & 2 & 0 \\
    Fairness/Cheating & 75 & \textbf{33} & 28 & 14 & 0\\
    Loyalty/Betrayal & 33 & \textbf{26} & 2 & 5 & 0 \\
    Authority/Subversion & 114 & 26 & \textbf{72} & 13 & 3 \\
    Purity/Degradation & 24 & 2 & \textbf{22} & 0 & 0 \\
    Liberty/Oppression & 93 & 9 & \textbf{78} & 6 & 0\\
    Non-moral & 304 & \textbf{188} & 68 & 44 & 4 \\ 
    No Agreement & 11 & 6 & 5 & 025 & 0\\
    \midrule
    \textsc{Total} & 750 & 367 & 292 & 84 & 7 \\
    \bottomrule 
    \end{tabular}}
    \caption{Dataset Summary}
    \label{tab:dataset}
\end{table}

\begin{figure*}[ht]
    \centering
\begin{subfigure}[ht]{0.7\columnwidth}
  \centering
  \includegraphics[width=\textwidth]{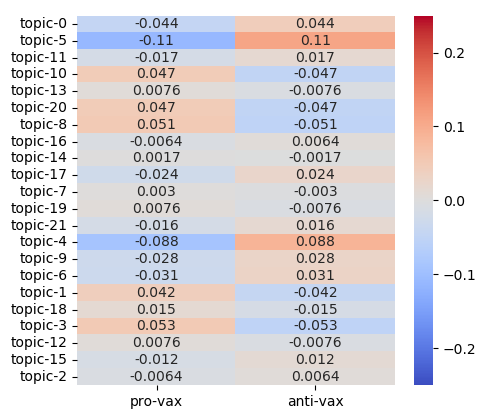}
  \caption{LDA Topics and Vax Stance}\label{fig:lda_stance}
\end{subfigure}
\begin{subfigure}{0.8\columnwidth}
  \centering
 \includegraphics[width=\textwidth]{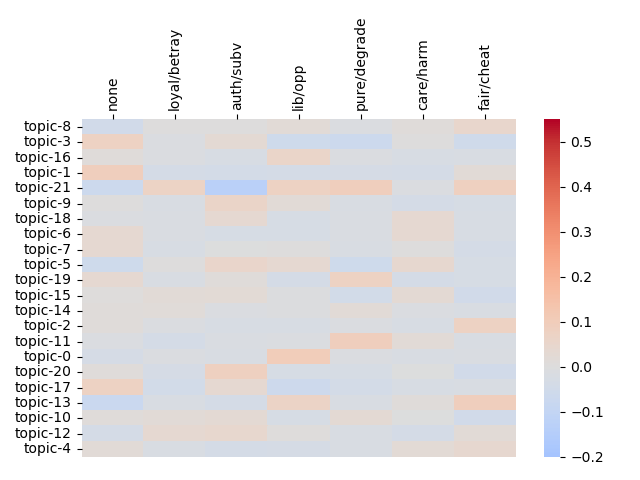}
    \caption{LDA Topics and Moral Foundations}\label{fig:lda_mf}
\end{subfigure}
\caption{Correlation Heatmaps for LDA Topics}\label{fig:lda}
\end{figure*}

To evaluate the correlation between the different dimensions of analysis, we calculate the Pearson correlation matrices and present them in Figure~\ref{fig:correl_comparison}. 
We can interpret reasons as distributions over moral foundations and stances (and vice-versa). This analysis provides a useful way to explain each of these dimensions. For example, we see that \textit{care/harm} is strongly correlated with reasons such as \textit{covid is real}, \textit{the vaccine works}, and \textit{natural immunity is ineffective}. Other expected trends emerge, such as \textit{purity/degradation} being highly correlated with \textit{vaccine against religion}. To evaluate the modeling advantage of our opinion analysis framework, we look at the correlation between stance, moral foundations and topics extracted in an unsupervised fashion using Latent Dirichlet Allocation (LDA)~\cite{blei2003latent}. We find that the reasons extracted interactively have higher correlations with both vaccination stance and moral foundations. The LDA correlation matrices can be seen in Figure \ref{fig:lda}.

In Table~\ref{tab:role_common} we show the top four reasons for \textit{fairness/cheating}. We choose this moral foundation given that it is evenly split among stances and it is active for different reasons. We show the top two (E,R,P) tuples for each reason. We can appreciate that while this moral foundation is used by people on both sides, the reasons offered and entities used vary. On the anti-vax side, authority figures and vaccine trials are portrayed as negative actors, while women and children are portrayed as targets. On the pro-vax side, Covid and unvaccinated people are portrayed as negative actors, and the general public is portrayed as a target.



\begin{table}[t]
\small
    \centering
    \resizebox{\columnwidth}{!}{%
    \begin{tabular}{c|c}
    \toprule
    \textbf{\textsc{VaxNotOppression}} & \textbf{\textsc{VaxDanger}} \\
    \midrule
    \textbf{70\% Pro-Vax} & \textbf{60\% Anti-Vax} \\
    (responsible people, target, neg)  & (pregnant women, target, neg) \\
    (un-vax people, actor, neg)  & (trial vax, actor, neg) \\
    \midrule
\textbf{\textsc{GovDistrust}} & \textbf{\textsc{VaxWorks}} \\
\midrule
\textbf{75\% Anti-Vax} & \textbf{75\% Pro-Vax}\\
    (children, target, neg) & (people, target, neg) \\
    (Fauci, actor, neg) & (COVID, actor, neg) \\
    \bottomrule
    \end{tabular}}\caption{Top 4 reasons for \textbf{Fairness/Cheating}, and their most frequent opinions and entity roles}
    \label{tab:role_common}
\end{table}


\begin{figure}[t]
    \centering
   
\end{figure}

\paragraph{Unlabeled Covid-19 vaccine corpus} In addition to our annotated dataset, we collected a corpus of 85,000 tweets in English mentioning the covid vaccine, uniformly distributed between January and October of 2021. These tweets are unlabeled, and are used to ground arguments (Section \ref{sec:opinions}) and to augment data for indirect supervision (Section \ref{sec:model}).

\section{Joint Probabilistic Model}\label{sec:model}

We propose a joint probabilistic model that reasons about the arguments made, their morality frames, stances, reasons, and the dependencies between them. We implement our model using DRaiL \cite{pacheco-goldwasser-2021-modeling}, a declarative modeling framework for specifying deep relational models. Deep relational models combine the strengths of deep neural networks and statistical relational learning (SRL) to model a joint distribution over relational data. This hybrid modeling paradigm allow us to leverage expressive textual encoders, and to introduce contextualizing information and model different interdependent decisions. SRL methods have proven effective to model domains with limited supervision \cite{johnson-goldwasser-2018-classification,subramanian-etal-2018-hierarchical}, and approaches that combine neural networks and SRL have shown consistent performance improvements \cite{widmoser-etal-2021-randomized, roy-etal-2021-identifying}. 



Following the conventions of statistical relational learning models, we use horn-clauses of the form $p_0 \wedge p_1 \wedge ... \wedge p_{n} \Rightarrow h$ to describe relational properties. Each logical rule defines a probabilistic scoring function over the relations expressed in its body and head. 


\paragraph{Base rules/classifiers} We define three base rules to score whether a tweet $\mathtt{t_i}$ has a moral judgment, what is its prominent moral foundation $\mathtt{m}$, and what is its vaccination stance.
\begin{equation}\label{eq:base_tweet}
\begin{split}
& r_0: \mathtt{Tweet(t_i)} \Rightarrow \mathtt{IsMoral(t_i)} \\
& r_1: \mathtt{Tweet(t_i)} \Rightarrow \mathtt{HasMF(t_i, m)} \\
& r_2: \mathtt{Tweet(t_i)} \Rightarrow \mathtt{VaxStance(t_i, s)} 
\end{split}
\end{equation}

To score the moral role of an entity $\mathtt{e_i}$ mentioned in tweet $\mathtt{t_i}$, we write two rules. The first one scores whether the entity $\mathtt{e_i}$ is an actor or a target, and the second one scores its polarity (positive or negative).
\begin{equation}\label{eq:base_entity}
\begin{split}
& r_3: \mathtt{Mentions(t_i, e_i)} \Rightarrow \mathtt{HasRole(e_i, r)} \\
& r_4: \mathtt{Mentions(t_i, e_i)} \Rightarrow \mathtt{EntPolarity(e_i, p)} 
\end{split}
\end{equation}

Note that these rules do not express any dependencies. They function as base classifiers that map tweets and entities to their most probable labels.


\paragraph{Dependency between roles and moral foundations} The way an entity is portrayed in a tweet can be highly indicative of its moral foundation. For example, people are likely to mention \textit{children} as a \textit{negative actor} in the context of \textit{care/harm}. To capture this, we explicitly model the dependency between an entity, its moral role, and the moral foundation.
\begin{equation}\label{eq:mf_role}
\begin{split}
& r_5: \mathtt{Mentions(t_i, e_j)} \wedge \mathtt{HasRole(e_i, r)} \\
& \;\;\; \wedge \mathtt{EntPolarity(e_i, p)} \Rightarrow \mathtt{HasMf(t_i, m)} 
\end{split}
\end{equation}

\paragraph{Dependency between stances and moral foundations} As we showed in Section \ref{sec:data}, there is a significant correlation between the stance of a tweet with respect to the vaccine debate, and its moral foundation. For example, people who oppose the vaccine are more likely to express the liberty/oppression moral foundation. To capture this, we model the dependency between the stance of a tweet and its moral foundation. 
\begin{equation}\label{eq:mf_stance}
\begin{split}
& r_6: \mathtt{VaxStance(t_i, s)} \Rightarrow \mathtt{HasMf(t_i, m)} 
\end{split}
\end{equation}

\paragraph{Dependency between reasons and moral foundations/stances} Explicitly modeling the dependency between repeating reasons and other decisions can help us add inductive bias into our model, potentially simplifying the task. For example, we can enforce the difference between two opposing views that use similar wording, and that could otherwise be treated similarly by a text-based model (e.g. \textit{``natural methods of protection against the disease are better than vaccines''} vs. \textit{`vaccines are better than natural methods of protection against the disease''}). We add two rules to capture this dependency, one between reasons and moral foundations, and one between reasons and stances.
\begin{equation}\label{eq:args}
\begin{split}
& r_7: \mathtt{Mentions(t_i, r)} \Rightarrow \mathtt{HasMf(t_i, m)} \\
& r_8: \mathtt{Mentions(t_i, r)} \Rightarrow \mathtt{VaxStance(t_i, s)} 
\end{split}
\end{equation}

\paragraph{Hard constraints} To enforce consistency between different decisions, we add two unweighted rules (or hard constraints). These rules are not associated with a scoring function and must always hold true. We enforce that, if a tweet is predicted to be moral, then it needs to also be associated to a specific moral foundation. Likewise, if a tweet is not moral, then no moral foundation should be assigned to it. 
\begin{equation}\label{eq:moral_mf}
\begin{split}
& c_0: \mathtt{IsMoral(t_i)} \Rightarrow \neg\mathtt{HasMf(t_i, none)}\\
& c_1: \neg\mathtt{IsMoral(t_i)} \Rightarrow \mathtt{HasMf(t_i, none)}\\
\end{split}
\end{equation}

Whenever different tweets have the same stance, we include a constraint to enforce consistency between the polarity of different mentions of the same entity. \citet{roy-etal-2021-identifying} showed that enforcing consistency for mentions of the same entity within a political party was beneficial. Given the polarization of the Covid-19 vaccine, we use the same rationale. 
\begin{equation}\label{eq:polarity}
\begin{split}
& c_3: \mathtt{Mentions(t_i, e_i)} \wedge \mathtt{Mentions(t_j, e_j)} \\
& \;\;\; \wedge \mathtt{SameStance(t_i, t_j)} \wedge \mathtt{EntPolarity(e_i, p)} \\
& \;\;\;\ \Rightarrow \mathtt{EntPolarity(e_j, p)}
\end{split}
\end{equation}

\paragraph{Learning and inference} The weights for each rule $w_r: p_0 \wedge p_1 \wedge ... \wedge p_{n} \Rightarrow h$ measure the importance of each rule in the model and can be learned from data. For example, when attempting to predict \textit{care/harm} for a tweet $\mathtt{t_i}$, we would like the weight of rule instance $\mathtt{IsTweet(t_i)} \Rightarrow \mathtt{HasMf(t_i, care/harm)}$ to be greater than the weight of rule instance $\mathtt{IsTweet(t_i)} \Rightarrow \mathtt{HasMf(t_i, loyalty/betrayal)}$. In DRaiL, these weights are learned using neural networks with parameters $\theta_r$. The collection of rules represents the global decision, and the solution is obtained by running a maximum a posteriori (MAP) inference procedure. Given that horn clauses can be expressed as linear inequalities corresponding to their disjunctive form, the MAP inference problem can be written as a linear program. DRaiL supports both locally and globally normalized structured prediction objectives. Throughout this paper, we used the locally normalized objective. For details about the learning procedure, we refer the reader to the original paper \cite{pacheco-goldwasser-2021-modeling}.

\paragraph{Learning with low-supervision} To learn DRaiL models in the low-supervision setting, we use an Expectation-Maximization style protocol, outlined in Algorithm \ref{alg:em}. First, we initialize the parameters of base rules using distant supervision classifiers. For moral foundations, we use the \citet{johnson-goldwasser-2018-classification} dataset and the Moral Foundation Twitter Corpus \cite{hoover2020mftc}. 
For roles, we use the \citet{roy-etal-2021-identifying} dataset. For polarity, we combine the \citet{roy-etal-2021-identifying} dataset with the MPQA 3.0 entity sentiment dataset~\cite{deng-wiebe-2015-mpqa}. For vaccination stances, we annotate our 85K unlabeled tweets using a set of prominent antivax and provax hashtags. Details about these datasets are provided in Appendix \ref{app:ood}.

Once the base rules have been initialized using distant supervision, we turn our attention to learning DRaiL models over the Covid-19 dataset presented in Section \ref{sec:data}. We alternate between MAP inference to obtain training labels (expectation step), and training the neural nets using these labels (maximization step). We receive an optional parameter $k$ indicating the amount of direct supervision to be used. When $k$ is provided, $k\%$ of the annotated labels are seeded during inference. 

\begin{algorithm}
\caption{\textit{Low Supervision Learning Protocol}}
\begin{algorithmic}[1]
\State Random initialization for all $\bm{\theta_r}$
\For{$r \in$ base rules} 
    \State $\bm{\theta_r} \gets $ distant supervision classifier
\EndFor
\While{not converged}
\State $\bm{Y_\mathtt{gold}} \gets $ DRaiL\_MAP\_inference(k)
\State Train all rules locally using $\bm{Y_\mathtt{gold}}$
\EndWhile
\end{algorithmic}
\label{alg:em}
\end{algorithm}
\begin{table*}[ht]
    \centering
    \resizebox{\textwidth}{!}{%
    \begin{tabular}{l|cc|cc|cc|cc|cc}
    \toprule
    \multirow{1}{*}{\textbf{\textsc{Model}}} & 
    \multicolumn{2}{c|}{\textbf{\textsc{Moral/NM}}} &
    \multicolumn{2}{c|}{\textbf{\textsc{Moral Found.}}} &
    \multicolumn{2}{c|}{\textbf{\textsc{Actor/Target}}} & \multicolumn{2}{c|}{\textbf{\textsc{Ent. Polarity}}} & \multicolumn{2}{c}{\textbf{\textsc{Vax Stance}}} \\
    ~ & Macro & Weighted & Macro & Weighted & Macro & Weighted & Macro & Weighted & Macro & Weighted \\
    \midrule
    Random & 54.96 & 55.36 & 11.07 & 15.15 & 45.57 & 45.72 & 34.63 & 36.69 & 49.16 & 49.23 \\
    Majority Class & 37.05 & 43.62 & 8.33 & 23.98 & 34.63 & 36.69 & 46.54 & 58.15 & 35.77 & 39.84 \\
    Lexicon Matching & 58.97 & 60.01 & 25.28 & 35.85 & - & - & - & - & - & - \\

    \midrule
    Base (distant sup.) & 69.77 & 68.88 & 28.79 & 41.27 & 71.94 & 72.05 & 63.88 & 74.30 & 69.46 & 70.35 \\
    Base (direct sup.) & 68.94 & 69.71 & 35.28 & 42.92 & \textbf{84.71} & \textbf{84.75} & \textbf{72.92} & \textbf{84.31} & 66.91 & 67.36 \\ 

    \textbf{+ Joint Model} & \textbf{80.53} & \textbf{81.17} &  \textbf{53.29} & \textbf{62.27} & 84.60 & 84.64 & 71.53 & 83.35 & \textbf{72.06} & \textbf{72.53} \\
    
    \bottomrule
    \end{tabular}
    }
    \caption{General Results (F1 Scores). NM: Non Moral}
    \label{tab:general_mf}
\end{table*}

\section{Experimental Evaluation}\label{sec:experiments}

The goal of our framework is to identify morality frames and opinions in tweets by modeling them jointly. In this section, we perform an exhaustive experimental analysis to evaluate the performance of our model and each of its components. 

\paragraph{Experimental settings} In DRaiL, each rule $r$ is associated with a neural architecture, which serves as a scoring function to obtain the rule weight $w_r$. We use BERT-base-uncased \cite{devlin2018bert} for all classifiers. For the rules that model dependencies (Eqs.~\ref{eq:mf_role},~\ref{eq:mf_stance},~\ref{eq:args}), we concatenate the CLS token with a one-hot vector of the symbols on the left hand side of the rule (i.e. role, sentiment, stance and reason), before passing it through a classifier. For rules that have the entity on the left-hand side (Eqs.~\ref{eq:base_entity},~\ref{eq:mf_role}), we use both the tweet and the entity as an input to BERT, using the SEP token. We trained supervised models using local normalization in DRaiL, and leveraged distant supervision using protocol outlined in Algorithm \ref{alg:em}. In all cases, we used a learning rate of $2e-5$, a maximum sequence length of $100$, and AdamW. In all experiments, we perform 5-fold cross-validation over the annotated dataset and report the micro-averaged results.

\paragraph{General results} Table~\ref{tab:general_mf} shows our general results for morality frames and vaccination stance. We evaluate our base classifiers and show the impact of modeling dependencies using DRaiL. The joint model results in a significant improvement for morality, moral foundation and vaccination stance. For entities, role and polarity remain stable. 
%
%
We also measure the impact of explicitly modeling reasons (Eq.~\ref{eq:args}) and present results in Table~\ref{tab:args}. We show the performance for the initial reasons proposed by~\citet{wawrzuta2021arguments}, which are all from the anti-vaccine perspective, and the impact of our two rounds of interaction, expanding and refining reasons (round 1) and augmenting argumentative patterns (round 2). We find that moral foundations improve from 60.07 to 62.27 and vaccination stance improves from 67.72 to 72.53 after interaction. 

\begin{table}[H]
    \centering
    \small
    \begin{tabular}{lcc}
    \toprule
    \textbf{\textsc{Model}} & \textbf{\textsc{MF}} & \textbf{\textsc{Vax. Stance}}  \\
    \midrule
        \textbf{ALL} (-Reasons) & 60.07 & 67.72\\
        + Reasons-Original & 61.51 & 72.62 \\
        + Reasons-Interaction-1 & 61.21 & \textbf{73.83} \\
        + Reasons-Interaction-2 &  \textbf{62.27} & 72.53 \\ 
    \bottomrule
    \end{tabular}
    \caption{Contribution of reasons at different interaction rounds (Weighted F1)}
    \label{tab:args}
\end{table}


\paragraph{Ablation study} We show an ablation study in Table~\ref{tab:ablation}. First, we can see how all dependencies contribute to the performance improvement, role-MF being the most impactful. We can also see that explicitly modeling morality constraints improves both the morality prediction and the moral foundation prediction, suggesting an advantage to breaking down this decision. We observe that the stance-polarity constraint does not have a significant impact, but does not hurt performance either, suggesting that our classifiers already capture this information. Lastly, we can see that the performance for roles and polarity remains stable, potentially because these classifiers have a strong starting point. 

\begin{table}[]
    \centering
    \resizebox{\columnwidth}{!}{%
    \begin{tabular}{lcccc}
    \toprule
    \multirow{1}{*}{\textbf{\textsc{Model}}} & 
    \multicolumn{1}{c}{\textbf{\textsc{M/NM}}} &
    \multicolumn{1}{c}{\textbf{\textsc{MF}}} &
    \multicolumn{1}{c}{\textbf{\textsc{Act/Tar}}} & \multicolumn{1}{c}{\textbf{\textsc{Polar.}}} \\ 
    \midrule
    
    \textbf{BERT} & 69.71 & 42.92 & \textbf{84.75} & 84.31 \\ 
    +RoleMF &  69.71 & 55.54 &  84.64 & 84.13 \\ 
    +RoleMF+MC & 79.00 & 57.68 & 84.64 & 84.13 \\
    +StanceMF & 69.71 & 47.85 & 84.75 & 84.31 \\
    +StanceMF+MC & 72.37 & 48.63 & 84.75 & 84.31 \\
    +StanceMF+MC+SPC & 72.32 & 48.63 & 84.75 & \textbf{84.35} \\
    +ReasonMF & 69.71  & 53.15 & 84.75 & 84.31 \\
    +ReasonMF+MC & 72.60 & 53.41 & 84.75 & 84.31 \\
    +ReasonStance+SPC & 69.71 & 42.92 & 84.64 & 83.26 \\
    \textbf{+ ALL} & \textbf{81.17} & \textbf{62.27} & 84.64 & 83.26 \\
    \bottomrule
    \end{tabular}}
    \caption{Ablation Study (Weighted F1). MC: Morality Constraint, SPC: Stance-Polarity Constraint}
    \label{tab:ablation}
\end{table}

\begin{figure}[t]
    \centering
    \includegraphics[width=\columnwidth]{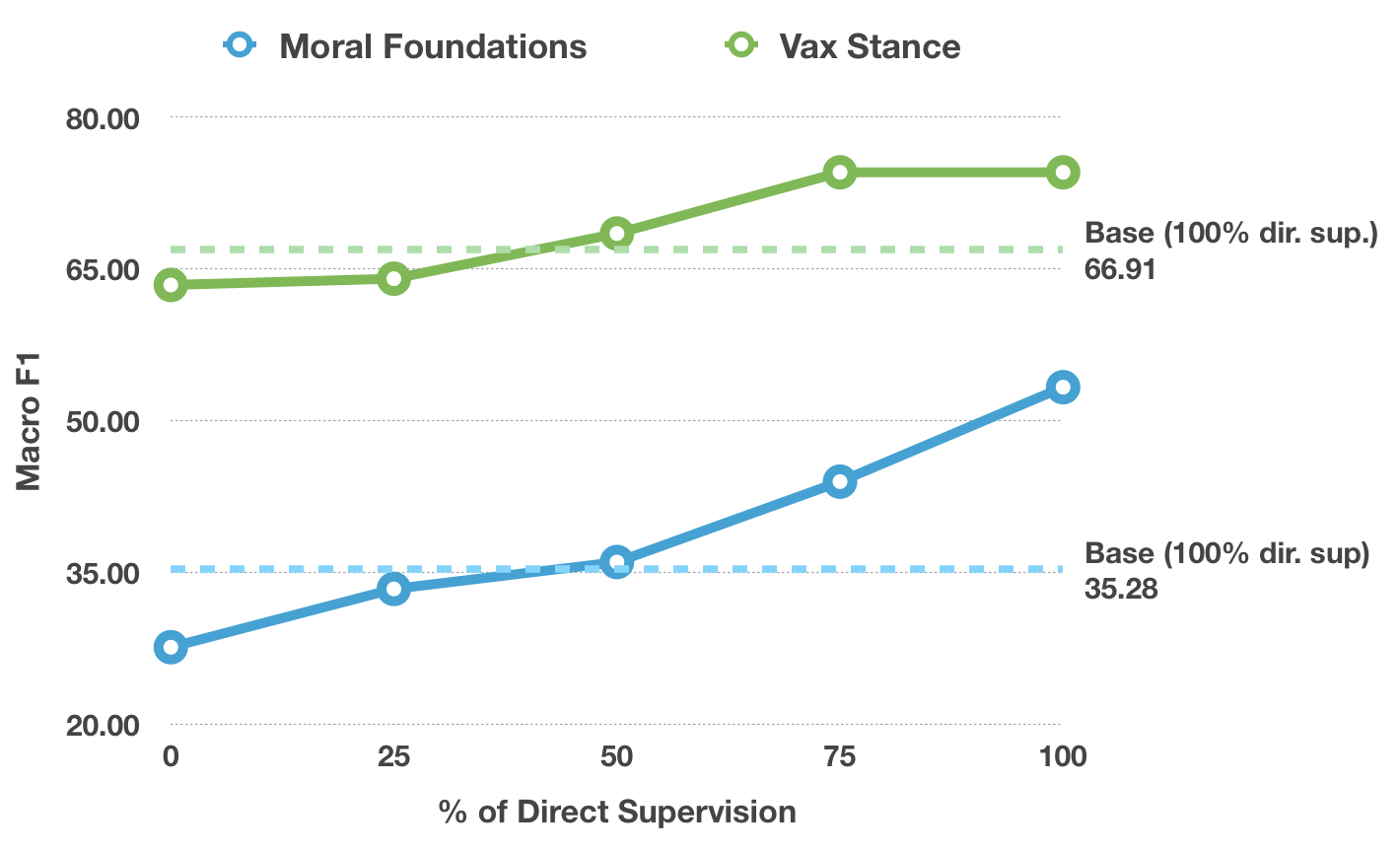}
    \caption{Performance in low-supervision settings}
    \label{fig:distant}
    \vspace{-10pt}
\end{figure}

\paragraph{Distant supervision} In Figure \ref{fig:distant} we evaluate the impact of our indirect supervision protocol by slowly augmenting the amount of direct supervision available. We can see that by leveraging out of domain-data and dependencies, we can obtain a competitive model using just 25\% of the annotated labels, and we can outperform the fully supervised classifiers using 50\% of the annotations.

\section{Related Work}

Recent studies have noted the prevalence of rumors and misinformation in the context of the Covid-19 pandemic~\cite{loomba2021measuring,shahi2021exploratory,lazarus2021global,ahmed2020covid}. Following this trend, several computational approaches have been proposed to detect misinformation related to Covid in news outlets and social media~\cite{weinzierl2021automatic,bang2021model,serrano2020nlp,al2020lies}. In this paper, we take a different approach and look at the problem of identifying opinions surrounding the Covid-19 vaccine, and explicitly modeling the rationale and moral sentiment that motivates them. 

Some recent works also look at analyzing arguments about Covid and vaccine hesitancy more broadly. In most cases, they either take a traditional classification approach for predicting stances~\cite{alliheibi2021opinion, lyu2021covid}, or use topic modeling techniques to uncover trends in word usage~\cite{skeppstedt2018vaccine,lyu2021covid, sha2020dynamic, zamani2020understanding}. In contrast, we propose a holistic framework that combines different methodological techniques, including human-in-the-loop mechanisms, classification with distant supervision, and deep relational learning to connect stance prediction, reason analysis and fine-grained entity moral sentiment analysis.


\section{Discussion}

We introduce a holistic framework for analyzing social media posts about the Covid-19 vaccine. We model morality frames and opinions jointly, and show that we can obtain competitive performance. The main limitation of our work is the size of the annotated dataset studied. Annotating for morality is a difficult and costly task, as it requires significant domain expertise. This motivates the need for methods that perform well under limited supervision, and that can leverage external and unlabeled resources. We took a first step in this direction by combining a wide range of methodological strategies. Given the amount of data generated daily about Covid, there are broader opportunities for exploiting these resources than what were explored in this paper. While we provided a preliminary analysis of the correlation between stances, reasons and morality, our current work looks at leveraging this framework to analyze opinions at scale. 

We also presented a first step towards interactive exploration of opinions on social media. While we explored this approach in a limited scenario, there is a lot of potential for using this paradigm for diagnosing NLP models and adapting to new domains. More research is required to devise protocols and evaluation strategies for this process.

\section{Acknowledgements}
We gratefully acknowledge our in-house annotators for their work annotating the dataset presented in this paper. We also thank the anonymous reviewers of this paper
for their insightful comments. This project was partially funded by a Microsoft Research Dissertation Grant, a Purdue Graduate School Summer Research Grant and an
NSF CAREER award IIS-2048001.

\bibliography{anthology,custom}

\begin{thebibliography}{43}
\expandafter\ifx\csname natexlab\endcsname\relax\def\natexlab#1{#1}\fi

\bibitem[{Ahmed et~al.(2020)Ahmed, Vidal-Alaball, Downing, Segu{\'\i}
  et~al.}]{ahmed2020covid}
Wasim Ahmed, Josep Vidal-Alaball, Joseph Downing, Francesc~L{\'o}pez
  Segu{\'\i}, et~al. 2020.
\newblock Covid-19 and the 5g conspiracy theory: social network analysis of
  twitter data.
\newblock \emph{Journal of medical internet research}, 22(5):e19458.

\bibitem[{Al-Rakhami and Al-Amri(2020)}]{al2020lies}
Mabrook~S Al-Rakhami and Atif~M Al-Amri. 2020.
\newblock Lies kill, facts save: detecting covid-19 misinformation in twitter.
\newblock \emph{Ieee Access}, 8:155961--155970.

\bibitem[{Alam et~al.(2021)Alam, Dalvi, Shaar, Durrani, Mubarak, Nikolov,
  Da~San~Martino, Abdelali, Sajjad, Darwish et~al.}]{alam2021fighting}
Firoj Alam, Fahim Dalvi, Shaden Shaar, Nadir Durrani, Hamdy Mubarak, Alex
  Nikolov, Giovanni Da~San~Martino, Ahmed Abdelali, Hassan Sajjad, Kareem
  Darwish, et~al. 2021.
\newblock Fighting the covid-19 infodemic in social media: A holistic
  perspective and a call to arms.
\newblock In \emph{Proceedings of the International AAAI Conference on Web and
  Social Media}, volume~15, pages 913--922.

\bibitem[{Alliheibi et~al.(2021)Alliheibi, Omar, and
  Al-Horais}]{alliheibi2021opinion}
Fahad~M Alliheibi, Abdulfattah Omar, and Nasser Al-Horais. 2021.
\newblock Opinion mining of saudi responses to covid-19 vaccines on twitter.
\newblock \emph{International Journal of Advanced Computer Science and
  Applications}, 12(6):72--78.

\bibitem[{Bang et~al.(2021)Bang, Ishii, Cahyawijaya, Ji, and
  Fung}]{bang2021model}
Yejin Bang, Etsuko Ishii, Samuel Cahyawijaya, Ziwei Ji, and Pascale Fung. 2021.
\newblock Model generalization on covid-19 fake news detection.
\newblock \emph{arXiv preprint arXiv:2101.03841}.

\bibitem[{Blei et~al.(2003)Blei, Ng, and Jordan}]{blei2003latent}
David~M Blei, Andrew~Y Ng, and Michael~I Jordan. 2003.
\newblock Latent dirichlet allocation.
\newblock \emph{Journal of machine Learning research}, 3(Jan):993--1022.

\bibitem[{Braun and Clarke(2012)}]{thematic_analysis}
Virginia Braun and Victoria Clarke. 2012.
\newblock \emph{Thematic analysis.}, pages 57--71.

\bibitem[{Chan(2021)}]{chan2021moral}
Eugene~Y Chan. 2021.
\newblock Moral foundations underlying behavioral compliance during the
  covid-19 pandemic.
\newblock \emph{Personality and individual differences}, 171:110463.

\bibitem[{Deng and Wiebe(2015)}]{deng-wiebe-2015-mpqa}
Lingjia Deng and Janyce Wiebe. 2015.
\newblock \href {https://doi.org/10.3115/v1/N15-1146} {{MPQA} 3.0: An
  entity/event-level sentiment corpus}.
\newblock In \emph{Proceedings of the 2015 Conference of the North {A}merican
  Chapter of the Association for Computational Linguistics: Human Language
  Technologies}, pages 1323--1328, Denver, Colorado. Association for
  Computational Linguistics.

\bibitem[{Devlin et~al.(2018)Devlin, Chang, Lee, and
  Toutanova}]{devlin2018bert}
Jacob Devlin, Ming-Wei Chang, Kenton Lee, and Kristina Toutanova. 2018.
\newblock Bert: Pre-training of deep bidirectional transformers for language
  understanding.
\newblock \emph{arXiv preprint arXiv:1810.04805}.

\bibitem[{D{\'\i}az and Cova(2021)}]{diaz2021reactance}
Rodrigo D{\'\i}az and Florian Cova. 2021.
\newblock Reactance, morality, and disgust: The relationship between affective
  dispositions and compliance with official health recommendations during the
  covid-19 pandemic.
\newblock \emph{Cognition and Emotion}, pages 1--17.

\bibitem[{Glandt et~al.(2021)Glandt, Khanal, Li, Caragea, and
  Caragea}]{glandt-etal-2021-stance}
Kyle Glandt, Sarthak Khanal, Yingjie Li, Doina Caragea, and Cornelia Caragea.
  2021.
\newblock \href {https://doi.org/10.18653/v1/2021.acl-long.127} {Stance
  detection in {COVID}-19 tweets}.
\newblock In \emph{Proceedings of the 59th Annual Meeting of the Association
  for Computational Linguistics and the 11th International Joint Conference on
  Natural Language Processing (Volume 1: Long Papers)}, pages 1596--1611,
  Online. Association for Computational Linguistics.

\bibitem[{Graham et~al.(2009)Graham, Haidt, and Nosek}]{graham2009}
Jesse Graham, Jonathan Haidt, and Brian~A Nosek. 2009.
\newblock Liberals and conservatives rely on different sets of moral
  foundations.
\newblock \emph{Journal of personality and social psychology}, 96(5):1029.

\bibitem[{Haidt and Graham(2007)}]{haidt2007morality}
Jonathan Haidt and Jesse Graham. 2007.
\newblock When morality opposes justice: Conservatives have moral intuitions
  that liberals may not recognize.
\newblock \emph{Social Justice Research}, 20(1):98--116.

\bibitem[{Hasan and Ng(2014)}]{hasan-ng-2014-taking}
Kazi~Saidul Hasan and Vincent Ng. 2014.
\newblock \href {https://doi.org/10.3115/v1/D14-1083} {Why are you taking this
  stance? identifying and classifying reasons in ideological debates}.
\newblock In \emph{Proceedings of the 2014 Conference on Empirical Methods in
  Natural Language Processing ({EMNLP})}, pages 751--762, Doha, Qatar.
  Association for Computational Linguistics.

\bibitem[{Hoover et~al.(2020)Hoover, Portillo-Wightman, Yeh, Havaldar, Davani,
  Lin, Kennedy, Atari, Kamel, Mendlen, Moreno, Park, Chang, Chin, Leong, Leung,
  Mirinjian, and Dehghani}]{hoover2020mftc}
J.~Hoover, G.~Portillo-Wightman, L.~Yeh, S.~Havaldar, A.M. Davani, Y.~Lin,
  B.~Kennedy, M.~Atari, Z.~Kamel, M.~Mendlen, G.~Moreno, C.~Park, T.E. Chang,
  J.~Chin, C.~Leong, J.Y. Leung, A.~Mirinjian, and M.~Dehghani. 2020.
\newblock Moral foundations twitter corpus: A collection of 35k tweets
  annotated for moral sentiment.
\newblock \emph{Social Psychological and Personality Science},
  11(8):1057--1071.

\bibitem[{Hossain et~al.(2020)Hossain, Logan~IV, Ugarte, Matsubara, Young, and
  Singh}]{hossain-etal-2020-covidlies}
Tamanna Hossain, Robert~L. Logan~IV, Arjuna Ugarte, Yoshitomo Matsubara, Sean
  Young, and Sameer Singh. 2020.
\newblock \href {https://doi.org/10.18653/v1/2020.nlpcovid19-2.11}
  {{COVIDL}ies: Detecting {COVID}-19 misinformation on social media}.
\newblock In \emph{Proceedings of the 1st Workshop on {NLP} for {COVID}-19
  (Part 2) at {EMNLP} 2020}, Online. Association for Computational Linguistics.

\bibitem[{Johnson and
  Goldwasser(2018)}]{johnson-goldwasser-2018-classification}
Kristen Johnson and Dan Goldwasser. 2018.
\newblock \href {https://doi.org/10.18653/v1/P18-1067} {Classification of moral
  foundations in microblog political discourse}.
\newblock In \emph{Proceedings of the 56th Annual Meeting of the Association
  for Computational Linguistics (Volume 1: Long Papers)}, pages 720--730,
  Melbourne, Australia. Association for Computational Linguistics.

\bibitem[{Krippendorff(2004)}]{krippendorff2004measuring}
Klaus Krippendorff. 2004.
\newblock Measuring the reliability of qualitative text analysis data.
\newblock \emph{Quality and quantity}, 38:787--800.

\bibitem[{Lazarus et~al.(2021)Lazarus, Ratzan, Palayew, Gostin, Larson, Rabin,
  Kimball, and El-Mohandes}]{lazarus2021global}
Jeffrey~V Lazarus, Scott~C Ratzan, Adam Palayew, Lawrence~O Gostin, Heidi~J
  Larson, Kenneth Rabin, Spencer Kimball, and Ayman El-Mohandes. 2021.
\newblock A global survey of potential acceptance of a covid-19 vaccine.
\newblock \emph{Nature medicine}, 27(2):225--228.

\bibitem[{Loomba et~al.(2021)Loomba, de~Figueiredo, Piatek, de~Graaf, and
  Larson}]{loomba2021measuring}
Sahil Loomba, Alexandre de~Figueiredo, Simon~J Piatek, Kristen de~Graaf, and
  Heidi~J Larson. 2021.
\newblock Measuring the impact of covid-19 vaccine misinformation on
  vaccination intent in the uk and usa.
\newblock \emph{Nature human behaviour}, 5(3):337--348.

\bibitem[{Lyu et~al.(2021)Lyu, Le~Han, and Luli}]{lyu2021covid}
Joanne~Chen Lyu, Eileen Le~Han, and Garving~K Luli. 2021.
\newblock Covid-19 vaccine--related discussion on twitter: topic modeling and
  sentiment analysis.
\newblock \emph{Journal of medical Internet research}, 23(6):e24435.

\bibitem[{Montagni et~al.(2021)Montagni, Ouazzani-Touhami, Mebarki, Texier,
  Sch{\"u}ck, Tzourio et~al.}]{montagni2021acceptance}
Ilaria Montagni, Kevin Ouazzani-Touhami, A~Mebarki, N~Texier, S~Sch{\"u}ck,
  Christophe Tzourio, et~al. 2021.
\newblock Acceptance of a covid-19 vaccine is associated with ability to detect
  fake news and health literacy.
\newblock \emph{Journal of public health (Oxford, England)}.

\bibitem[{Muric et~al.(2021)Muric, Wu, and Ferrara}]{Muric2021}
Goran Muric, Yusong Wu, and Emilio Ferrara. 2021.
\newblock \href {http://arxiv.org/abs/2105.05134} {{COVID-19 Vaccine Hesitancy
  on Social Media: Building a Public Twitter Dataset of Anti-vaccine Content,
  Vaccine Misinformation and Conspiracies}}.
\newblock \emph{arxiv}.

\bibitem[{Pacheco and Goldwasser(2021)}]{pacheco-goldwasser-2021-modeling}
Maria~Leonor Pacheco and Dan Goldwasser. 2021.
\newblock \href {https://doi.org/10.1162/tacl_a_00357} {Modeling content and
  context with deep relational learning}.
\newblock \emph{Transactions of the Association for Computational Linguistics},
  9:100--119.

\bibitem[{Pagliaro et~al.(2021)Pagliaro, Sacchi, Pacilli, Brambilla, Lionetti,
  Bettache, Bianchi, Biella, Bonnot, Boza et~al.}]{pagliaro2021trust}
Stefano Pagliaro, Simona Sacchi, Maria~Giuseppina Pacilli, Marco Brambilla,
  Francesca Lionetti, Karim Bettache, Mauro Bianchi, Marco Biella, Virginie
  Bonnot, Mihaela Boza, et~al. 2021.
\newblock Trust predicts covid-19 prescribed and discretionary behavioral
  intentions in 23 countries.
\newblock \emph{PloS one}, 16(3):e0248334.

\bibitem[{Reimers and Gurevych(2019)}]{reimers-gurevych-2019-sentence}
Nils Reimers and Iryna Gurevych. 2019.
\newblock \href {https://doi.org/10.18653/v1/D19-1410} {Sentence-{BERT}:
  Sentence embeddings using {S}iamese {BERT}-networks}.
\newblock In \emph{Proceedings of the 2019 Conference on Empirical Methods in
  Natural Language Processing and the 9th International Joint Conference on
  Natural Language Processing (EMNLP-IJCNLP)}, pages 3982--3992, Hong Kong,
  China. Association for Computational Linguistics.

\bibitem[{Rousseeuw(1987)}]{Rousseeuw87silhouetteCluster}
Peter Rousseeuw. 1987.
\newblock \href
  {https://doi.org/http://dx.doi.org/10.1016/0377-0427(87)90125-7}
  {Silhouettes: a graphical aid to the interpretation and validation of cluster
  analysis}.
\newblock \emph{J. Comput. Appl. Math.}, 20(1):53--65.

\bibitem[{Roy et~al.(2021)Roy, Pacheco, and
  Goldwasser}]{roy-etal-2021-identifying}
Shamik Roy, Maria~Leonor Pacheco, and Dan Goldwasser. 2021.
\newblock \href {https://aclanthology.org/2021.emnlp-main.783} {Identifying
  morality frames in political tweets using relational learning}.
\newblock In \emph{Proceedings of the 2021 Conference on Empirical Methods in
  Natural Language Processing}, pages 9939--9958, Online and Punta Cana,
  Dominican Republic. Association for Computational Linguistics.

\bibitem[{Serrano et~al.(2020)Serrano, Papakyriakopoulos, and
  Hegelich}]{serrano2020nlp}
Juan Carlos~Medina Serrano, Orestis Papakyriakopoulos, and Simon Hegelich.
  2020.
\newblock Nlp-based feature extraction for the detection of covid-19
  misinformation videos on youtube.
\newblock In \emph{Proceedings of the 1st Workshop on NLP for COVID-19 at ACL
  2020}.

\bibitem[{Sha et~al.(2020)Sha, Hasan, Mohler, and Brantingham}]{sha2020dynamic}
Hao Sha, Mohammad~Al Hasan, George Mohler, and P~Jeffrey Brantingham. 2020.
\newblock Dynamic topic modeling of the covid-19 twitter narrative among us
  governors and cabinet executives.
\newblock \emph{arXiv preprint arXiv:2004.11692}.

\bibitem[{Shahi et~al.(2021)Shahi, Dirkson, and
  Majchrzak}]{shahi2021exploratory}
Gautam~Kishore Shahi, Anne Dirkson, and Tim~A Majchrzak. 2021.
\newblock An exploratory study of covid-19 misinformation on twitter.
\newblock \emph{Online social networks and media}, 22:100104.

\bibitem[{Skeppstedt et~al.(2018)Skeppstedt, Kerren, and
  Stede}]{skeppstedt2018vaccine}
Maria Skeppstedt, Andreas Kerren, and Manfred Stede. 2018.
\newblock Vaccine hesitancy in discussion forums: computer-assisted argument
  mining with topic models.
\newblock In \emph{Building Continents of Knowledge in Oceans of Data: The
  Future of Co-Created eHealth}, pages 366--370. IOS Press.

\bibitem[{Sowa et~al.(2021)Sowa, Kiszkiel, Laskowski, Alimowski,
  Szczerbi{\'n}ski, Paniczko, Moniuszko-Malinowska, and
  Kami{\'n}ski}]{sowa2021covid}
Pawe{\l} Sowa, {\L}ukasz Kiszkiel, Piotr~Pawe{\l} Laskowski, Maciej Alimowski,
  {\L}ukasz Szczerbi{\'n}ski, Marlena Paniczko, Anna Moniuszko-Malinowska, and
  Karol Kami{\'n}ski. 2021.
\newblock Covid-19 vaccine hesitancy in poland—multifactorial impact
  trajectories.
\newblock \emph{Vaccines}, 9(8):876.

\bibitem[{Subramanian et~al.(2018)Subramanian, Cohn, and
  Baldwin}]{subramanian-etal-2018-hierarchical}
Shivashankar Subramanian, Trevor Cohn, and Timothy Baldwin. 2018.
\newblock \href {https://doi.org/10.18653/v1/N18-1178} {Hierarchical structured
  model for fine-to-coarse manifesto text analysis}.
\newblock In \emph{Proceedings of the 2018 Conference of the North {A}merican
  Chapter of the Association for Computational Linguistics: Human Language
  Technologies, Volume 1 (Long Papers)}, pages 1964--1974, New Orleans,
  Louisiana. Association for Computational Linguistics.

\bibitem[{Tagliabue et~al.(2020)Tagliabue, Galassi, and
  Mariani}]{tagliabue2020pandemic}
Fabio Tagliabue, Luca Galassi, and Pierpaolo Mariani. 2020.
\newblock The “pandemic” of disinformation in covid-19.
\newblock \emph{SN comprehensive clinical medicine}, 2(9):1287--1289.

\bibitem[{van~der Maaten and Hinton(2008)}]{vanDerMaaten2008}
Laurens van~der Maaten and Geoffrey Hinton. 2008.
\newblock \href {http://www.jmlr.org/papers/v9/vandermaaten08a.html}
  {Visualizing data using {t-SNE}}.
\newblock \emph{Journal of Machine Learning Research}, 9:2579--2605.

\bibitem[{Wawrzuta et~al.(2021)Wawrzuta, Jaworski, Gotlib, and
  Panczyk}]{wawrzuta2021arguments}
Dominik Wawrzuta, Mariusz Jaworski, Joanna Gotlib, and Mariusz Panczyk. 2021.
\newblock What arguments against covid-19 vaccines run on facebook in poland:
  Content analysis of comments.
\newblock \emph{Vaccines}, 9(5):481.

\bibitem[{Weinzierl et~al.(2021)Weinzierl, Hopfer, and
  Harabagiu}]{weinzierl2021misinformation}
Maxwell Weinzierl, Suellen Hopfer, and Sanda~M Harabagiu. 2021.
\newblock Misinformation adoption or rejection in the era of covid-19.
\newblock In \emph{Proceedings of the International AAAI Conference on Web and
  Social Media}, volume~15, pages 787--795.

\bibitem[{Weinzierl and Harabagiu(2021)}]{weinzierl2021automatic}
Maxwell~A Weinzierl and Sanda~M Harabagiu. 2021.
\newblock Automatic detection of covid-19 vaccine misinformation with graph
  link prediction.
\newblock \emph{Journal of biomedical informatics}, 124:103955.

\bibitem[{Widmoser et~al.(2021)Widmoser, Pacheco, Honorio, and
  Goldwasser}]{widmoser-etal-2021-randomized}
Manuel Widmoser, Maria~Leonor Pacheco, Jean Honorio, and Dan Goldwasser. 2021.
\newblock \href {https://doi.org/10.18653/v1/2021.eacl-main.100} {Randomized
  deep structured prediction for discourse-level processing}.
\newblock In \emph{Proceedings of the 16th Conference of the European Chapter
  of the Association for Computational Linguistics: Main Volume}, pages
  1174--1184, Online. Association for Computational Linguistics.

\bibitem[{Zamani et~al.(2020{\natexlab{a}})Zamani, Schwartz, Eichstaedt,
  Guntuku, Ganesan, Clouston, and Giorgi}]{zamani2020understanding}
Mohammadzaman Zamani, H~Andrew Schwartz, Johannes Eichstaedt, Sharath~Chandra
  Guntuku, Adithya~Virinchipuram Ganesan, Sean Clouston, and Salvatore Giorgi.
  2020{\natexlab{a}}.
\newblock Understanding weekly covid-19 concerns through dynamic
  content-specific lda topic modeling.
\newblock In \emph{Proceedings of the Conference on Empirical Methods in
  Natural Language Processing. Conference on Empirical Methods in Natural
  Language Processing}, volume 2020, page 193. NIH Public Access.

\bibitem[{Zamani et~al.(2020{\natexlab{b}})Zamani, Schwartz, Eichstaedt,
  Guntuku, Virinchipuram~Ganesan, Clouston, and
  Giorgi}]{zamani-etal-2020-understanding}
Mohammadzaman Zamani, H.~Andrew Schwartz, Johannes Eichstaedt, Sharath~Chandra
  Guntuku, Adithya Virinchipuram~Ganesan, Sean Clouston, and Salvatore Giorgi.
  2020{\natexlab{b}}.
\newblock \href {https://doi.org/10.18653/v1/2020.nlpcss-1.21} {Understanding
  weekly {COVID}-19 concerns through dynamic content-specific {LDA} topic
  modeling}.
\newblock In \emph{Proceedings of the Fourth Workshop on Natural Language
  Processing and Computational Social Science}, pages 193--198, Online.
  Association for Computational Linguistics.

\end{thebibliography}
\bibliographystyle{acl_natbib}
\appendix

\section{Appendix}

\subsection{Reasons and Phrases}\label{app:themes_phrases}
\begin{figure*}[htbp]
  \centering  
  \includegraphics[width=  1 \textwidth]{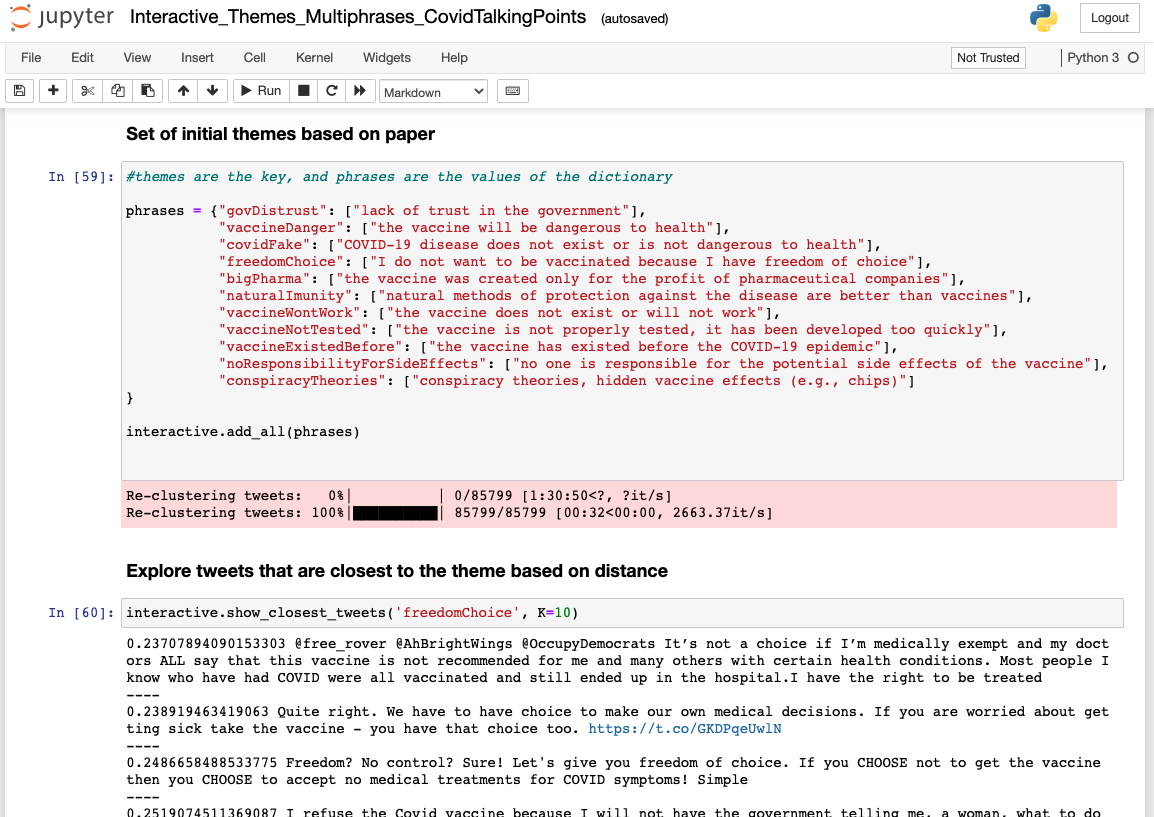}
    \caption{Interactive task interface.}
    \label{fig:inter1}
\end{figure*}

\begin{figure*}[htbp]
  \centering  
  \includegraphics[width=  1 \textwidth]{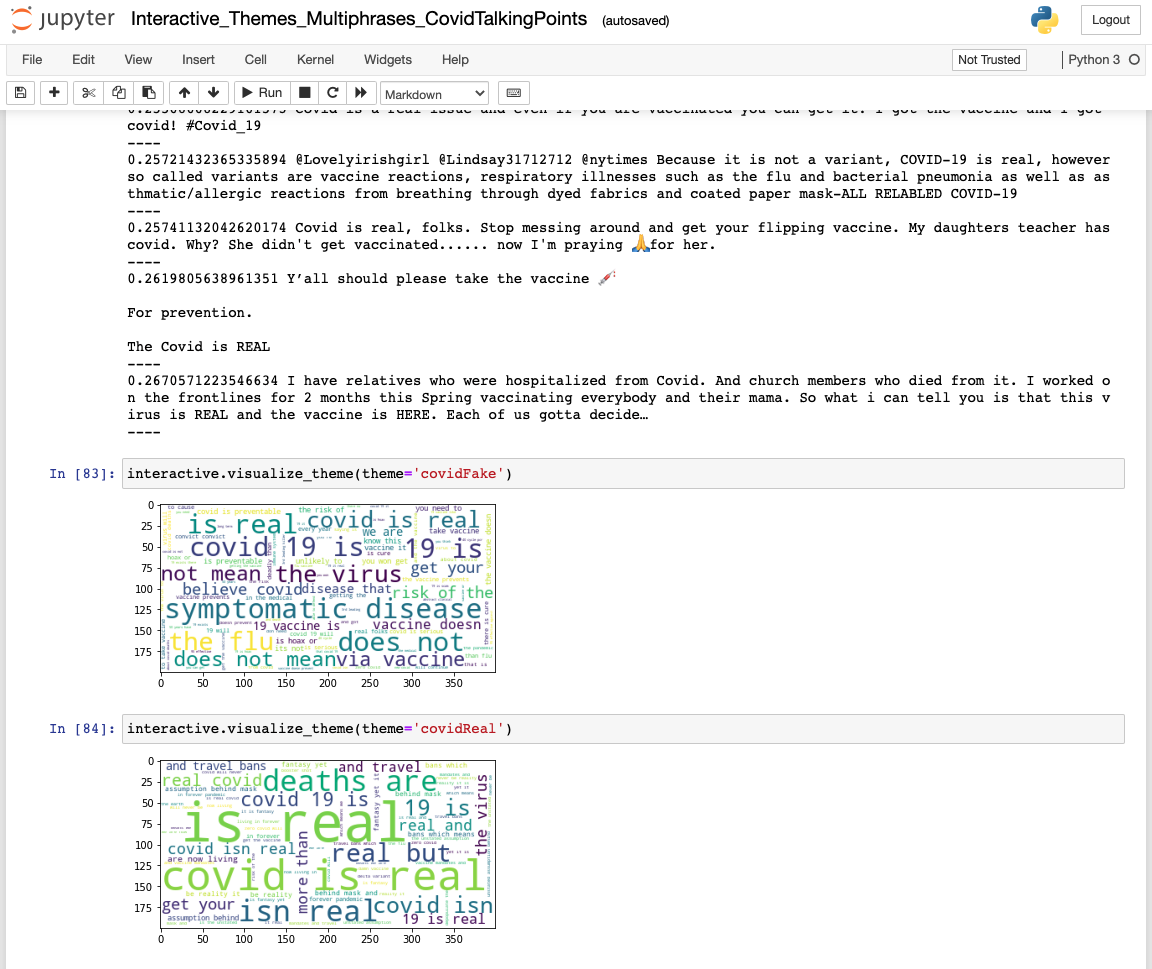}
    \caption{After querying the themes (i.e., CovidFake, CovidReal), interface shows the wordcloud.}
    \label{fig:inter2}
\end{figure*}

\begin{figure*}
\begin{subfigure}{\columnwidth}
  \centering
  \includegraphics[width=\textwidth]{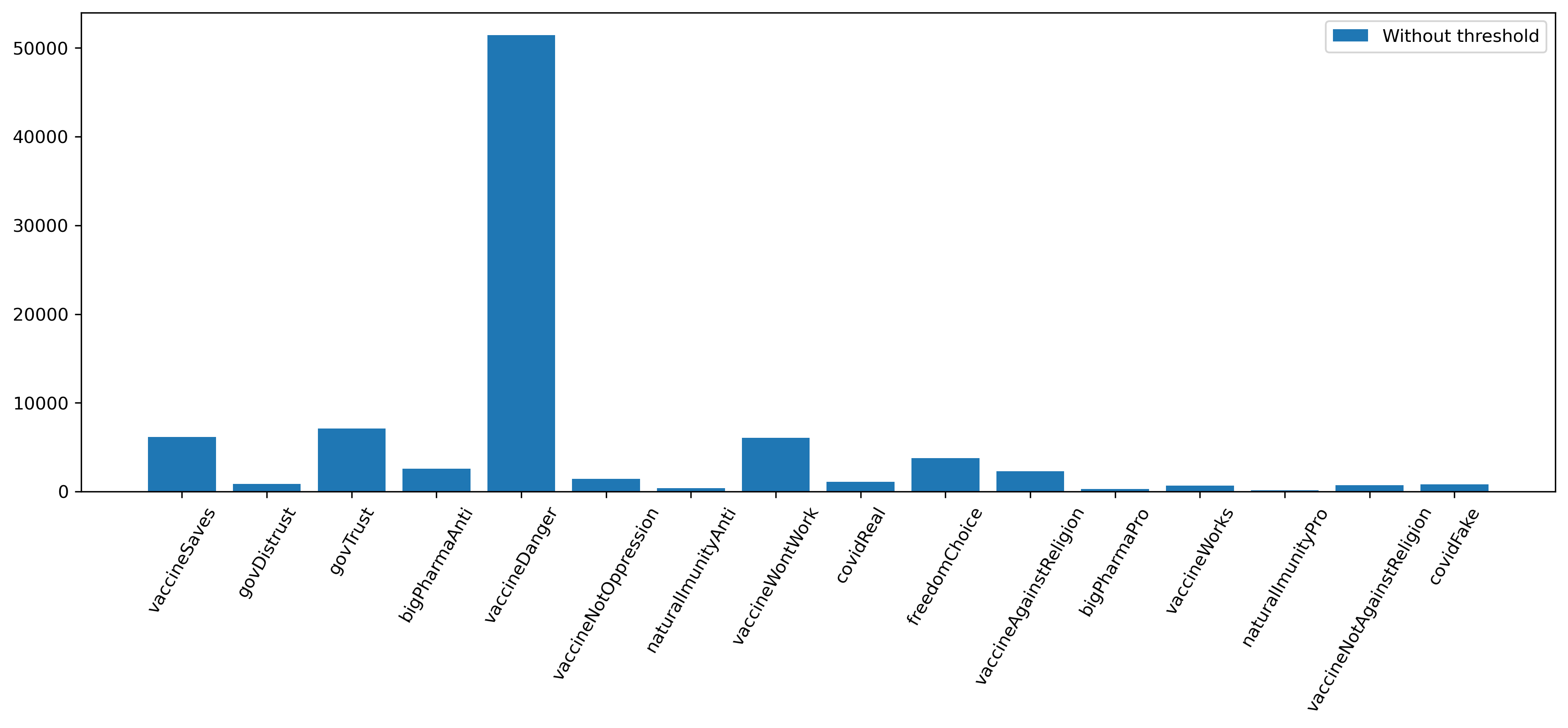}
  \caption{Without threshold before}
  \label{fig:Without_threshold_before}
\end{subfigure}%
\begin{subfigure}{\columnwidth}
  \centering
  \includegraphics[width=\textwidth]{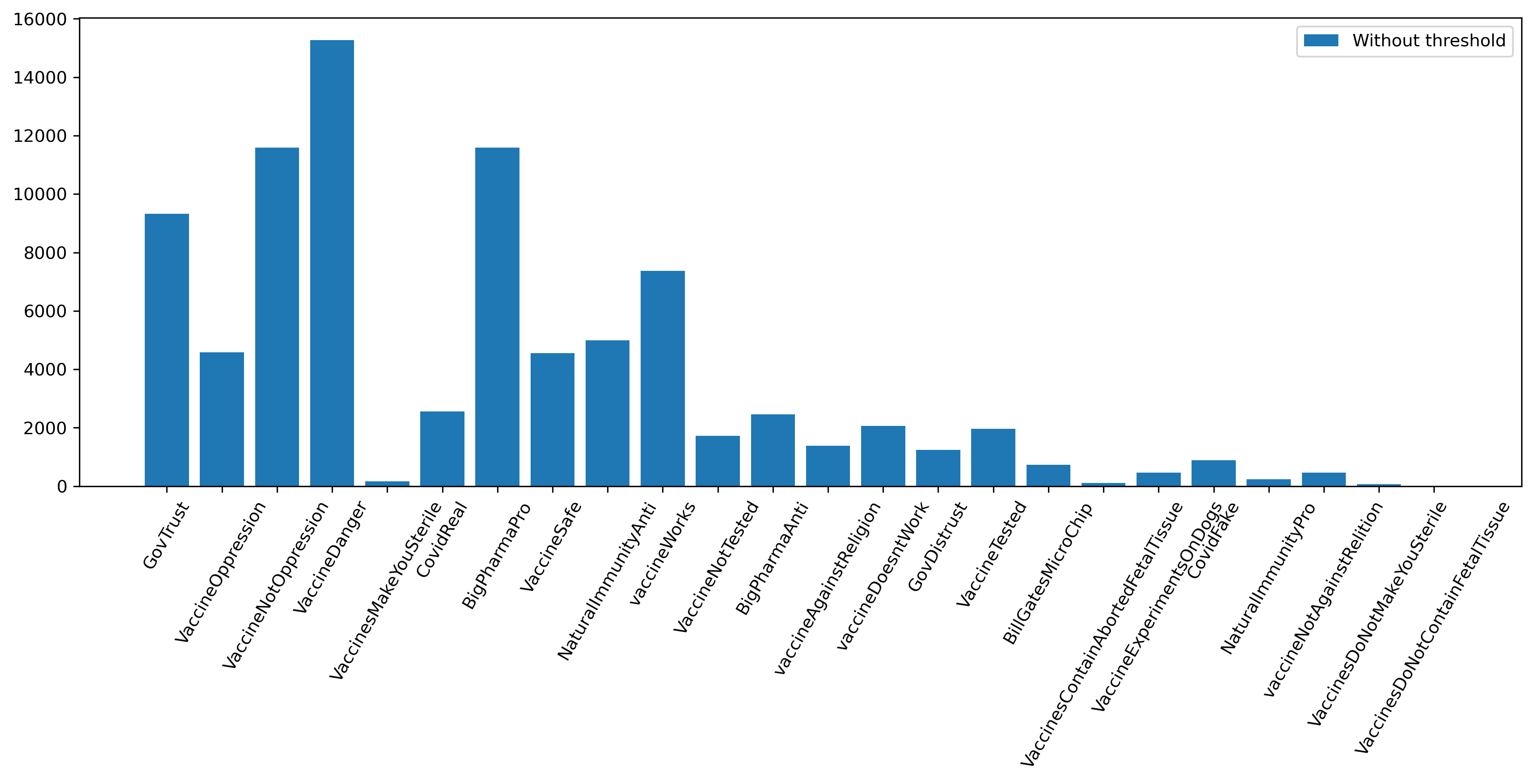}
  \caption{Without threshold after }
  \label{fig:Without_threshold_after}
\end{subfigure}
\begin{subfigure}{\columnwidth}
  \centering
  \includegraphics[width=\textwidth]{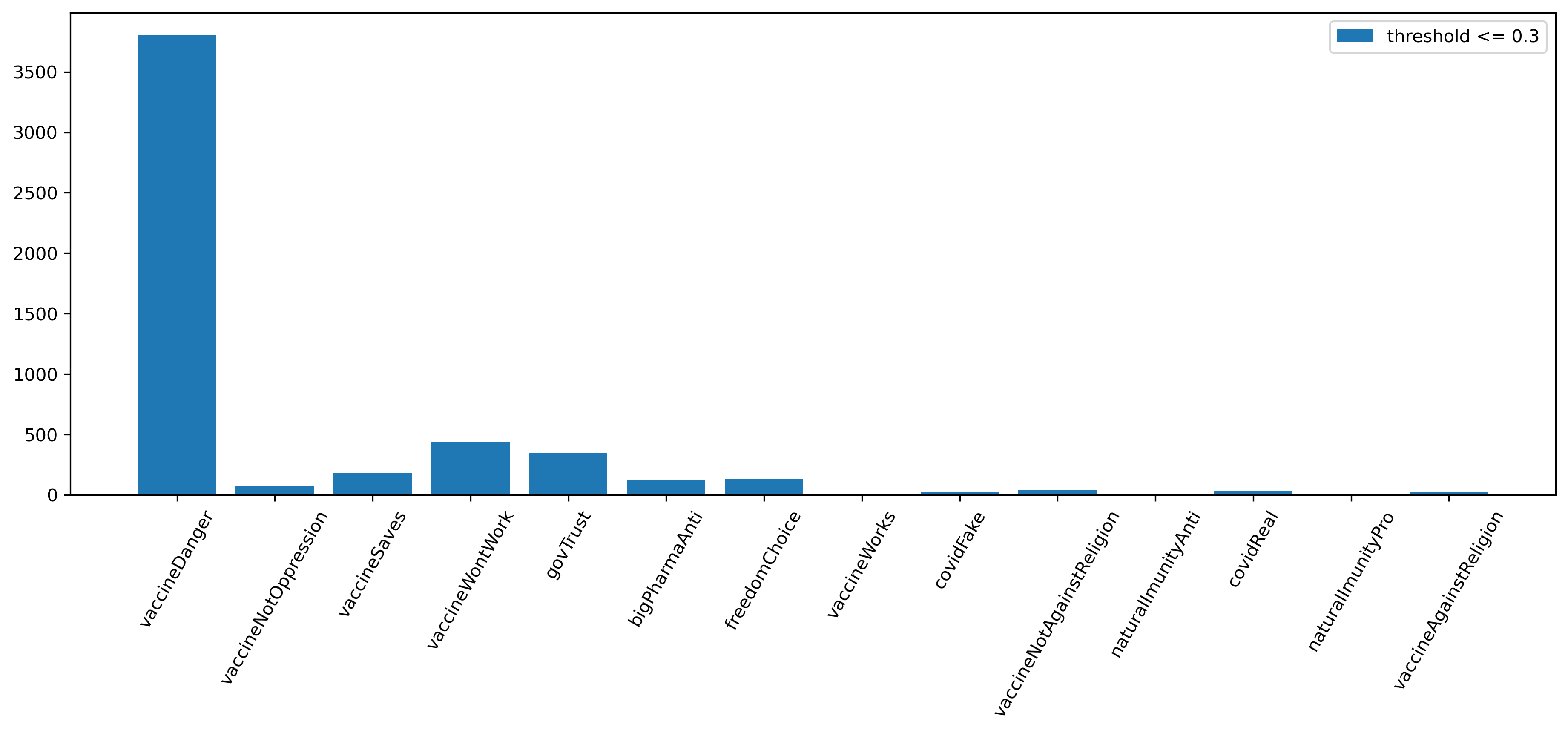}
  \caption{$threshold \leq 0.3 $ before}
  \label{fig:thr_0.3_before}
\end{subfigure}%
\begin{subfigure}{\columnwidth}
  \centering
  \includegraphics[width=\textwidth]{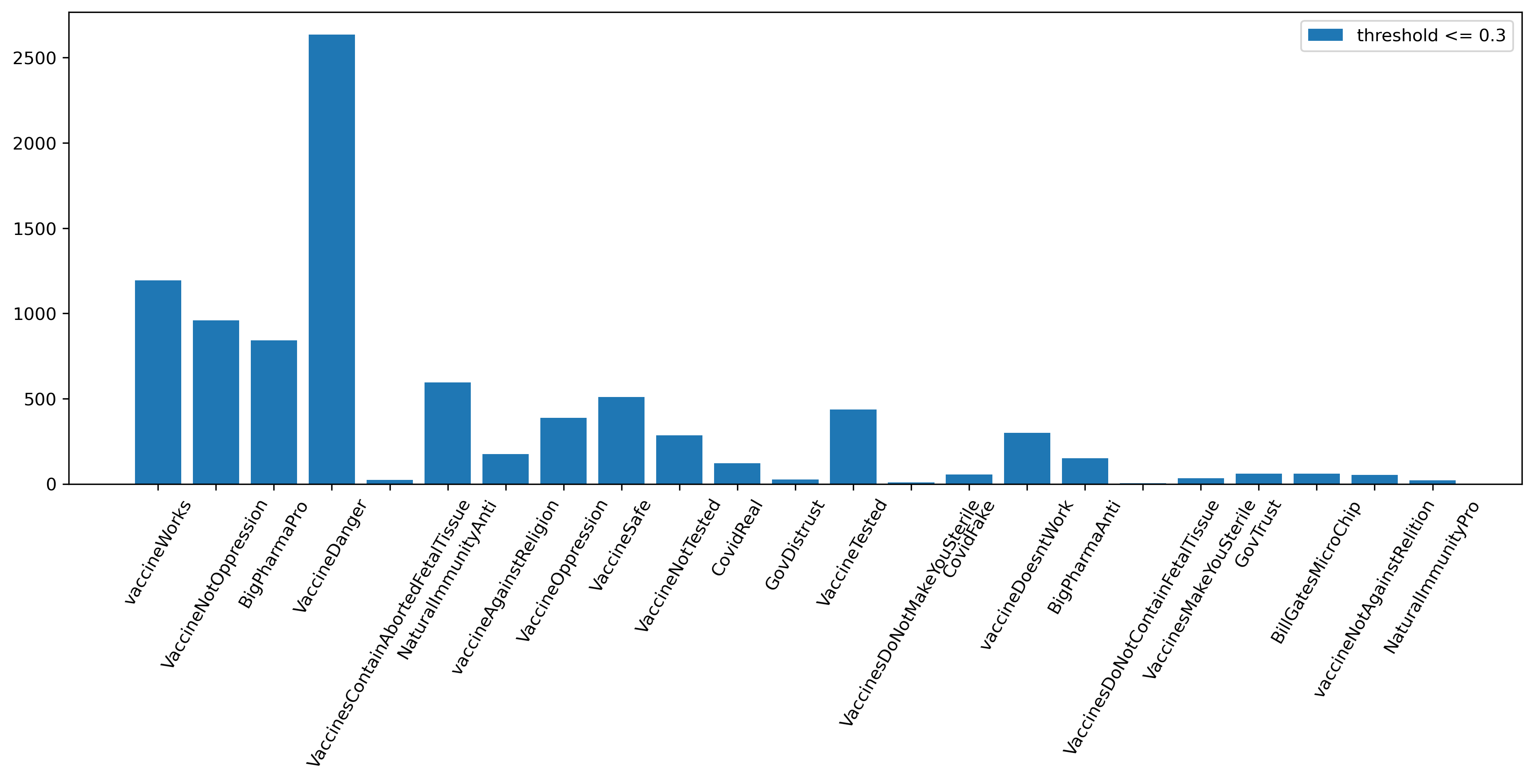}
  \caption{$threshold \leq 0.3 $ after}
  \label{fig:thr_0.3_after}
\end{subfigure}
\caption{Cluster assignment before and after refining arguments interactively. }
\label{fig:bar}
\end{figure*}

Tables \ref{tab:themesAnti} and \ref{tab:themesPro} show the full list of phrases for anti-vax and pro-vax reasons. The interactive task interface is presented in Figures \ref{fig:inter1} and \ref{fig:inter2}. Bar plots for reason assignments before and after interaction are shown in Figure \ref{fig:bar}.

\begin{figure}[H]
\begin{subfigure}{.5\columnwidth}
  \centering
  \includegraphics[width=\textwidth]{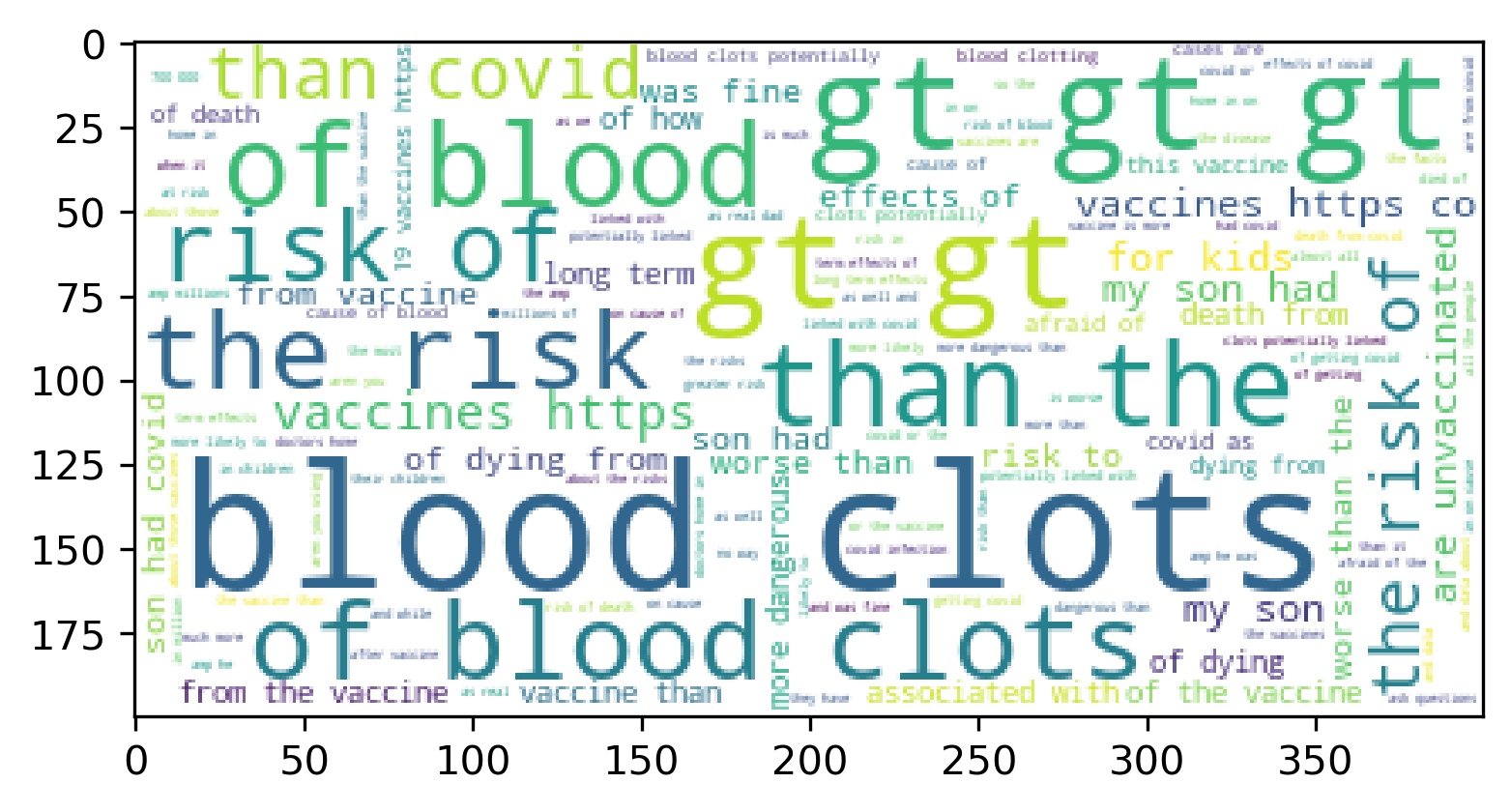}
  \caption{Theme: VaccineDanger}
  \label{fig:VaccineDanger_before}
\end{subfigure}%
\begin{subfigure}{.5\columnwidth}
  \centering
  \includegraphics[width=\textwidth]{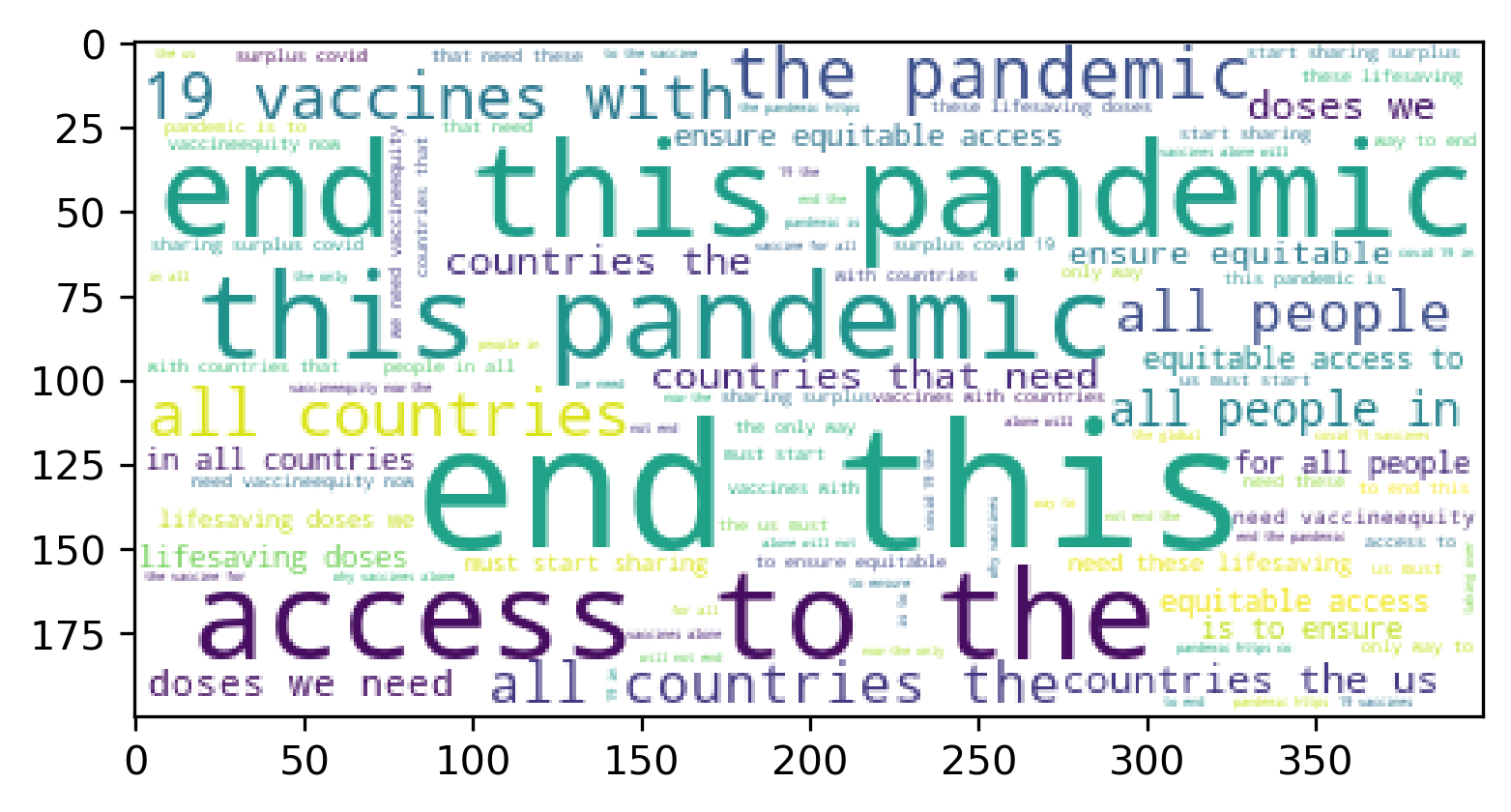}
  \caption{Theme: VaccineSafe}
  \label{fig:VaccineSafe_before}
\end{subfigure}
\caption{Wordclouds for reasons \textbf{before} interaction.}
\label{fig:wc_before}
\end{figure}
\vspace{-5mm}
\begin{figure}[H]
\begin{subfigure}{.5\columnwidth}
  \centering
  \includegraphics[width=\textwidth]{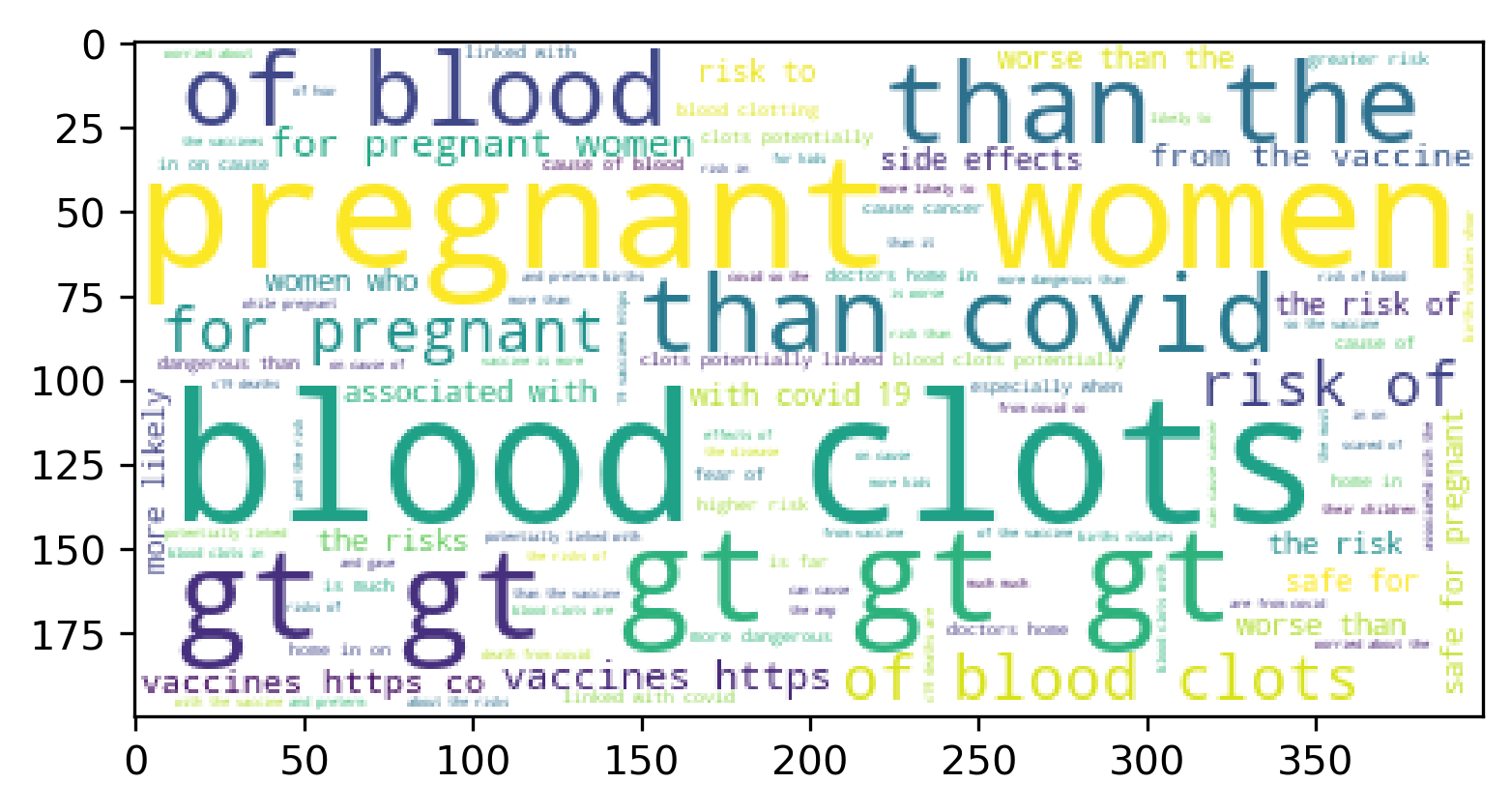}
  \caption{Theme: VaccineDanger}
  \label{fig:VaccineDanger_after}
\end{subfigure}%
\begin{subfigure}{.5\columnwidth}
  \centering
  \includegraphics[width=\textwidth]{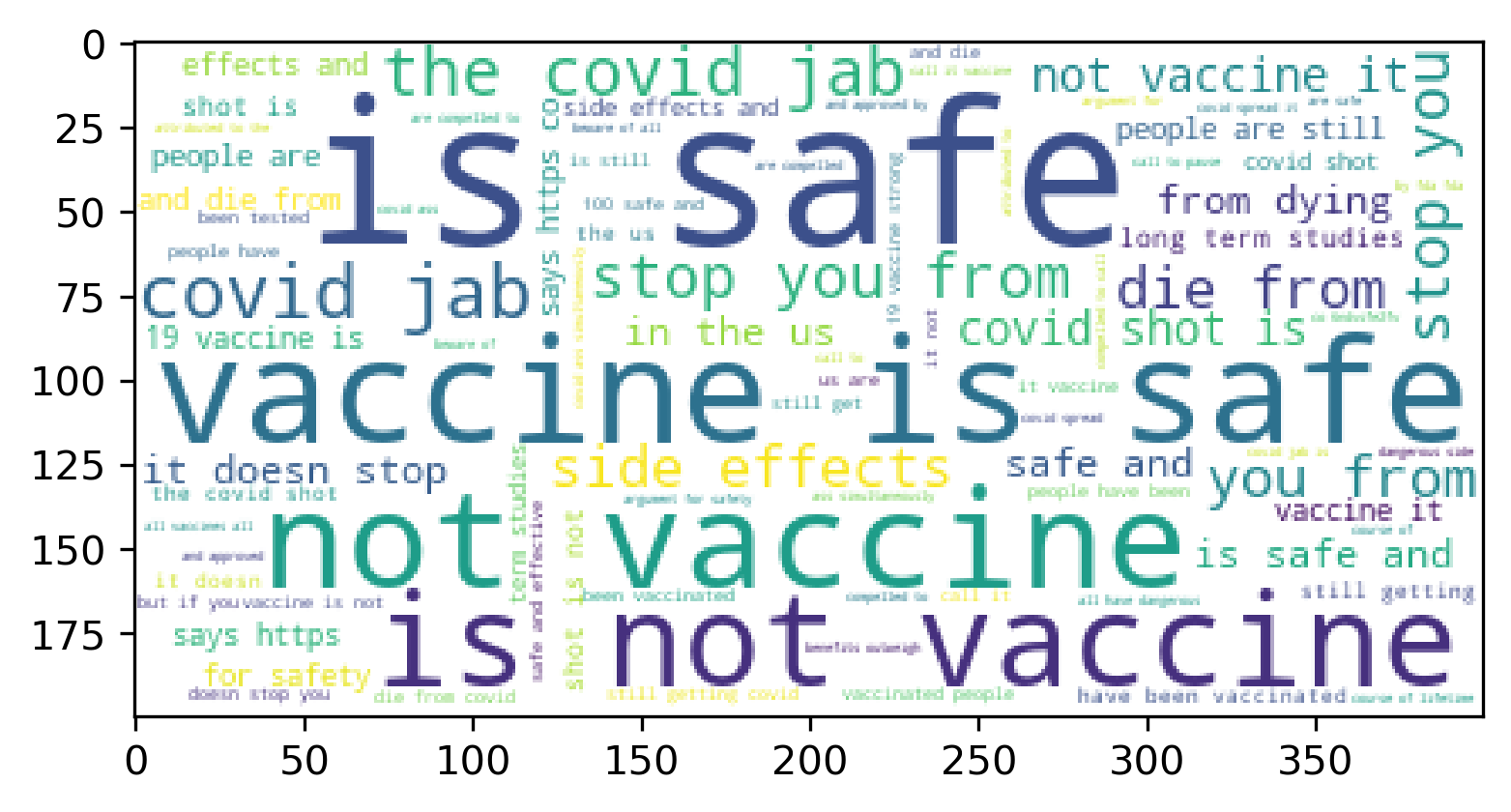}
  \caption{Theme: VaccineSafe}
  \label{fig:VaccineSafe_after}
\end{subfigure}
\caption{Wordclouds for reasons \textbf{after} interaction. }
\label{fig:wc_after}
\end{figure}

\begin{table*}[]
    \centering
    \resizebox{\textwidth}{!}{%
    \begin{tabular}{|l|l|}
      \hline
      \thead{Themes} & \thead{Overarching Patterns} \\
      \hline
      \textbf{GovDistrust} & Add phrases with strong word for distrust \\
      ~ & “Good at being bad” \\
      ~ & Explicit negations \\
      \hline 
      \textbf{GovTrust} & Hedging phrases (sort-of trust) \\
      \hline
      \textbf{VaxDanger} &  Closer connection between vaccine words and danger words (related to sickness, bad effects) \\
      & Explit negations \\
      & Rhetorical questions \\
      & Refusing the vaccine for medical reasons \\
      \hline
      \textbf{VaxSafe} & Explicit mentions of safety \\
      & Explicit negations \\
      \hline
      \textbf{CovidFake} & Stronger relevant negative words (fake, scam, hoax) \\
      & Explicit negations \\
      \hline
      \textbf{CovidReal} & Trust the science \\
      & References to Covid hospitalization on the rise, explicit mentions of hospitals \\
      & Explicit negations \\
      \hline
      \textbf{VaxOppression} & Legal language \\
      & Explicit mentions of discrimination and oppression \\
      & Sarcasm \\
      \hline
      \textbf{VaxNotOppression} & Justifying mandates \\
      & Freedom to be protected \\
      & Criticizing others using “you/people” language, focus freedom on me/my/I \\
      \hline
      \textbf{BigPharmaAnti} & Stronger words against pharmaceutical companies (corrupt, evil) \\
      & Not accountable / irresponsible past behavior \\
      & Mentions of negative side-effect of other products (cancer) \\
      \hline
     \textbf{BigPharmaPro} & Trust science/research and vaccine development process \\
     & Language about intent, the vaccine was created to do something good, explicit names of companies \\
     \hline
     \textbf{NaturalImmunityPro} & The vaccine is not enough \\
     & Explicit mentions to population immunity, herd immunity and antibodies \\
     \hline
     \textbf{NaturalImmunityAnti} & Emphasis on global look, collective entities, society \\
     & Natural immunity characterized as dangerous or not effective \\
     & Mentions of experts and trusting science \\
     \hline 
     \textbf{VaxAgainstReligion} & I put it in god hands (god is deciding) \\
     & Treating pro-vax as another religion \\
     \hline
     \textbf{VaxNotAgainstReligion} & “Religious” in quotes \\
     & Bugus exemptions \\
     & “Where is your faith” \\
     & Call to action: get tested/get vaccinated/put a mask on (mentions of compassion) \\
     & No religion ask members to refuse vaccine \\
     \hline
     \textbf{VaxDoesntWork} & Reference to “magic vaccine” \\
     & “Never developed”, “doesn’t work” \\
     & Questions: why are deaths high? Why is corona not going away? Why are vaccinated people dying?
    \\
     \hline
     \textbf{VaxWorks} & “ask a doctor”, consult with an expert \\
     & Research on the vaccine is good/has been going on for a long time \\
     & Capture differences, e.g. “good trials” vs. rushed ones. \\
     \hline
     \textbf{VaxNotTested} & Language suggesting “rushed through trials” and “experimental vaccine” \\
     \hline
     \textbf{VaxTested} & trust the research and development process \\
     & Testing can be confused with covid-test, use other language.
    \\
     \hline

    \end{tabular}}
    \caption{Overarching argumentation patterns uncovered by coders during interaction}
    \label{tab:arg_patterns}
\end{table*}

\begin{table*}
\centering
    \resizebox{\textwidth}{!}{%
    \begin{tabular}{|l|l|}
      \hline
      \thead{Themes} & \thead{Phrases} \\
      \hline
      \textbf{GovDistrust} &  \makecell[l]{\textbf{"lack of trust in the government"},
    "Fuck the government", 
    "The government is a total failure", \\
    "Never trust the government", 
    "Biden is a failure", 
    "Biden lied people die", \\
    "The government and Fauci have been dishonest",
    "The government always lies",\\
    "The government has a strong record of screwing things up",
    "The government is good at screwing things up",\\
    "The government is screwing things up",
    "The government is lying",
    "The government only cares about money", \\
    "The government doesn't work logically", 
    "Do not trust the government", \\
    "The government doesn't care about people’s health",
    "The government won't tell you the truth about the vaccine"}    \\
      \hline
   
      \textbf{VaxDanger} &  \makecell[l]{\textbf{"the vaccine will be dangerous to health"},
    "Covid vaccines can cause blood clots",\\
    "The vaccine is a greater danger to our children’s health than COVID itself",\\
    "The vaccine will kill you",
    "The experimental covid vaccine is a death jab",\\
    "The covid vaccine causes cancer",
    "The covid vaccine is harmful for pregnant women and kids",\\
    "The vaccine increases health risk",
    "The vaccine isn't safe", \\
    "What are vaccines good for? Nothing, rather it increases risk",\\
    "I and many others have medical exemptions",
    "The vaccine is dangerous for people with medical conditions",\\
    "I won't take the vaccine due to medical reasons",
    "The vaccine has dangerous side effects"}    \\
      \hline
      
     \textbf{CovidFake} &  \makecell[l]{\textbf{"Covid-19 disease does not exist"},
    "Covid is fake",
    "covid is a hoax",
    "covid is a scam",\\
    "covid is propaganda",
    "the pandemic is a lie",
    "covid isn't real", 
    "I don't think that covid is real", \\
    "I don't buy that covid is real", 
    "I don't think there is a pandemic", \\
    "I don't think the pandemic is real",
    "I don't buy that there is a pandemic"}    \\
      \hline
      \textbf{VaxOppression} &  \makecell[l]{
     \textbf{"I do not want to be vaccinated because I have freedom of choice"}\\
     "Forcing people to take experimental vaccines is oppression", \\
    "The vaccine has nothing to do with Covid-19, it's about the vaccine passport and tyranny",\\
    "The vaccine mandate is unconstitutional",
     "I choose not to take the vaccine", \\
     "My body my choice",
    "I'm not against the vaccine but I am against the mandate", \\
    "I have freedom to choose not to take the vaccine",
    "I am free to refuse the vaccine", \\
    "It is not about covid, it is about control",
    "Medical segregation based on vaccine mandates is discrimination", \\
    "The vaccine mandate violates my rights",
    "Falsely labeling the injection as a vaccine is illegal", \\
    "Firing over vaccine mandates is oppression",
    "Vaccine passports are medical tyranny", \\
    "I won't let the government tell me what I should do with my body",
    "I won't have the government tell me what to do"}    \\

     \hline
     \textbf{BigPharmaAnti} &  \makecell[l]{\textbf{"the vaccine was created only for the profit of pharmaceutical companies"},\\
     "We are the subjects of massive experiments for the Moderna and Pfizer vaccines", \\
    "Pharmaceutical companies are corrupt",
    "The pharmaceutical industry is rotten",
    "Big Pharma is evil", \\
    "How would you trust big pharma with the COVID vaccine? They haven’t been liable for vaccine harm in the past", \\
    "Covid vaccines are not doing what the pharmaceutical companies promised", \\
    "Pharmaceutical companies have a history of irresponsible behavior", \\
    "I don't trust Johnson \& Johnson after knowing their baby powder caused cancer for decades"}    \\
     \hline

      \textbf{NatImmunityPro} &  \makecell[l]{\textbf{"natural methods of protection against the disease are better than vaccines"}, \\
    "Herd immunity is broad, protective, and durable", \\
    "Natural immunity has higher level of protection than the vaccine",
    "Embrace population immunity", \\
    "I trust my immune system",
    "I have antibodies I do not need the vaccine",
    "Natural immunity is effective"}    \\
     \hline

     \textbf{\textcolor{blue}{VaxAgainstReligion}} &  \makecell[l]{"The vaccine is against my religion",
    "The vaccines are the mark of the beast",
    "The vaccine is a tool of Satan", \\
    "The vaccine is haram",
    "The vaccine is not halal",\\
    "I will protect my body from a man made vaccine",
    "I put it all in God's hands",
    "God will decide our fate", \\
    "The vaccine contains bovine, which conflicts with my religion", \\
    "The vaccine contains aborted fetal tissue which is against my religion", \\
    "The vaccine contains pork, muslims can't take the vaccine",
    "Jesus will protect me", \\
    "The vaccine doesn't protect you from getting or spreading Covid, God does",
    "The covid vaccine is another religion"}    \\
     \hline

     \textbf{VaxDoesntWork} &  \makecell[l]{\textbf{"the vaccine does not work"},
    "covid vaccines do not stop the spread", \\
    "If the vaccine works, why are deaths so high?",
    "Why are vaccinated people dying?", \\
    "If the vaccine works, why is covid not going away?"}    \\
     \hline
    \textbf{VaxNotTested} &  \makecell[l]{\textbf{"the vaccine is not properly tested, it has been developed too quickly"},\\
    "Covid-19 vaccines have not been through the same rigorous testing as other vaccines", \\
    "The Covid vaccine is experimental",
    "The covid vaccine was rushed through trials", \\
    "The approval of the experimental vaccine was rushed",
    "How was the vaccine developed so quickly?"}    \\
    \hline
    \textbf{\textcolor{blue}{VaxExperimentDogs}} &  \makecell[l]{"Animal shelters are empty because Dr Fauci allowed \\ experimenting of various Covid vaccines/drugs on dogs and other domestic pets", \\
    "Fauci tortures dogs and puppies"}    \\
    \hline
    \textbf{\textcolor{blue}{BillGatesMicroChip}} &  \makecell[l]{"The covid vaccine is a ploy to microchip people", \\
    "Bill Gates wants to use vaccines to  implant microchips in people", \\
    "Globalists support a covert mass chip implantation through the covid vaccine"}    \\
    \hline
    \textbf{\textcolor{blue}{VaxFetalTissue}} &  \makecell[l]{"There is aborted fetal tissue in the Covid Vaccines",
    "the Covid vaccines contain aborted fetal cells"}    \\
    
    \hline
    \textbf{\textcolor{blue}{VaxMakeYouSterile}} &  \makecell[l]{"The covid vaccine will make you sterile",
    "Covid vaccine will affect your fertility"}    \\
    \hline
    \textbf{\textcolor{red}{NoResponsibility}} & \textbf{no one is responsible for the potential side effects of the vaccine} \\
    \hline
    \textbf{\textcolor{red}{SwineFluVax}} & \textbf{mentioning the past development of the swine flu vaccine} \\
    \hline
    \textbf{\textcolor{red}{VaxResistance}} & \textbf{the vaccine has existed before the Covid-19 epidemic, now there is too much resistance} \\
    \hline 
    \textbf{\textcolor{red}{ConspiracyTheories}} & \textbf{conspiracy theories, hidden vaccine effects (e.g., chips)} \\
    \hline
    
    \end{tabular}}
    \caption{AntiVax Themes and phrases for Covid-19 talking points. Themes that were added during interaction are shown in blue. Themes that were removed during interaction are shown in red. The original explanations/examples are presented in bold. }
\label{tab:themesAnti}
\end{table*}

\begin{table*}
\centering
    \resizebox{\textwidth}{!}{%
    \begin{tabular}{|l|l|}
    \hline
      \thead{Themes} & \thead{Phrases} \\
    \hline
       \textbf{\textcolor{blue}{GovTrust}} &  \makecell[l]{"We trust the government",
    "The government cares for people", \\
    "We are thankful to the government for the vaccine availability", \\
    "Hats off to the government for tackling the pandemic",\\
    "It is a good thing to be skeptical of the government, but they are right about the covid vaccine",\\
    "It is a good thing to be skeptical of the government, but they haven’t lied about the covid vaccine", \\
    "The government can be corrupt, but they are telling the truth about the covid vaccine", \\
    "The government can be corrupt, but they are not lying about the covid vaccine"}    \\
      \hline
      \textbf{\textcolor{blue}{VaxSafe}} &  \makecell[l]{"The vaccine is safe",
    "Millions have been vaccinated with only mild side effects", \\
    "Millions have been safely vaccinated against covid",
    "The benefits of the vaccine outweigh its risks", \\
    "The vaccine has benefits",
    "The vaccine is safe for women and kids",
    "The vaccine won't make you sick",\\
    "The vaccine isn't dangerous",
    "The vaccine won't kill you",\\
    "The covid vaccine isn't a death jab",
    "The covid vaccine doesn't harm women and kids"}    \\
      \hline
     \textbf{\textcolor{blue}{CovidReal}} &  \makecell[l]{"Covid is real",
    "I trust science",
    "Covid death is real", \\
    "The science doesn't lie about covid",
    "Scientist know what they are doing", \\
    "Scientist know what they are saying",
    "Covid hospitalizations are on the rise", \\
    "Covid hospitalizations are climbing as fourth stage surge continues", \\
    "Covid's death toll has grown faster",
    "Covid is not a hoax",
    "The pandemic is not a lie",\\
    "The pandemic is not a lie, hospitalizations are on the rise"}    \\
    \hline
    \textbf{\textcolor{blue}{VaxNotOppression}} &  \makecell[l]{"The vaccine mandate is not oppression because vaccines lower hospitalizations and death rates", \\
    "The vaccine mandate is not oppression because it will help to end this pandemic", \\
    "The vaccine mandate will help us end the pandemic", \\
    "We need a vaccine mandate to end this pandemic",
    "I support vaccine mandates", \\
    "If you don't get the vaccine based on your freedom of choice, \\ don’t come crawling to the emergency room when you get COVID", \\
    "If you refuse a free FDA-approved vaccine for non-medical reasons, \\ 
    then the government shouldn't continue to give you free COVID tests", \\
    "You are free not to take the vaccine, businesses are also free to deny you entry", \\
    "You are free not to take the vaccine, businesses are free to protect their customers and employees", \\
    "If you choose not to take the vaccine, you have to deal with the consequences", \\
    "If it is your body your choice, then insurance companies should stop paying for your hospitalization costs for COVID"}    \\
    \hline
    \textbf{\textcolor{blue}{BigPharmaPro}} &  \makecell[l]{"I trust the science and pharmaceutical research",
    "Pharmaceutical companies are not hiding anything", \\
    "The research behind covid vaccines is public",
    "The Pfizer vaccine is saving lives", \\
    "The Moderna vaccines are helping stop the spread of covid", \\
    "The Johnson and Johnson vaccine was created to stop covid", \\
    "Pharmaceutical companies are seeking FDA approval",
    "Pharmaceutical companies are following standard protocols"}    \\
     \hline
    \textbf{\textcolor{blue}{NatImmunityAnti}} &  \makecell[l]{"Only the vaccine will end the pandemic", \\
    "Vaccines will allow us to defeat covid without death and sickness", \\
    "The vaccine has better long term protection than to natural immunity",
    "Natural immunity is not effective",\\
    "Natural immunity would require a lot of people getting sick", \\
    "Experts recommend the vaccine over natural immunity"}    \\
     \hline
    \textbf{\textcolor{blue}{VaxReligionOk}} &  \makecell[l]{"The vaccine is not against religion, get the vaccine", 
    "No religion ask members to refuse the vaccine", \\
    "Religious exemptions are bogus", \\
    "When turning in your religious exemption forms for the vaccine, remember ignorance is not a religion", \\
    "Disregard for others' lives isn't part of your religion", \\
    "Jesus is trying to protect us from covid by divinely inspiring scientists to create vaccines"}    \\
     \hline
    \textbf{\textcolor{blue}{VaxWorks}} &  \makecell[l]{"The vaccine works",
    "Vaccines do work, ask a doctor or consult with an expert", \\
    "The covid vaccine helps to stop the spread",
    "Unvaccinated people are dying at a rapid rate from Covid-19", \\
    "There is a lot of research supporting that vaccines work", \\
    "The research on the covid vaccine has been going on for a long time"}    \\
     \hline
    \textbf{\textcolor{blue}{VaxTested}} &  \makecell[l]{"Covid vaccine research has been going on for a while",
    "Plenty of research has been done on the covid vaccine", \\
    "The technologies used to develop the Covid-19 vaccines \\ have been in development for years to prepare for outbreaks of infectious viruses",\\
    "The testing processes for the vaccines were thorough didn't skip any steps",
    "The vaccine received FDA approval"}    \\
    \hline
    \textbf{\textcolor{red}{ProVax}} & \textbf{positive attitude} \\
    \hline
    
\end{tabular}}
    \caption{ProVax reasons and phrases. Reasons that were added during interaction are shown in blue. Reasons that were removed during interaction are shown in red. The original explanatory phrases are presented in bold.}
\label{tab:themesPro}
\end{table*}

\subsection{Data Collection}\label{app:data_collection}

To create the list of keywords used to collect tweets about the Covid-19 vaccine, we read multiple articles about Covid mentioning vaccination status, vaccine hesitancy, misinformation, vaccine constraints, health issues, religious sentiment and other vaccine-related debates, and made a list of repeating statements. Then, we consulted three researchers, two in Computational Social Science and one in Psychology, and constructed a list of relevant keywords that are indicative of morally charged discussions. The full list of keywords can observed in Table \ref{tab:keywords}.

\begin{table}[H]
\begin{adjustbox}{width=\columnwidth, center}
\begin{tabular}{|l|}
\hline
covid vaccine,
covid vaccination,
covid vaccine tyranny, \\
covid vaccine oppression,
covid vaccine mandate,
covid vaccine conspiracy,\\
covid vaccine anti-vax,
covid vaccine religion,
covid vaccine satan,\\
covid vaccine god,
covid vaccine jesus,
covid vaccine islam, \\
covid vaccine muslim,
covid vaccine christianity,
covid vaccine christian, \\
covid vaccine hindu,
covid vaccine jews,
covid vaccine catholic,\\
covid vaccine buddhism,
covid vaccine religious,
covid vaccine biden failure,\\
covid vaccine passport,
covid vaccine loyalty,
covid vaccine cheating,\\
covid vaccine freedom,
covid vaccine betrayal,
covid vaccine liberty,\\
covid vaccine black people,
covid vaccine propaganda,
covid vaccine hesitancy,\\
covid vaccine hesitant,
covid vaccine microchip,
covid vaccine bill,\\
covid vaccine pregnancy,
covid vaccine pregnant,
covid vaccine approval,\\
covid vaccine biden,
covid vaccine fda,
covid vaccine cdc,\\
covid vaccine fauci,
Covid-19 china,
vaccine passport,\\
vaccination mandate,
covid vaccine death,
covid vaccine military,\\
experimental covid vaccine,
covid vaccine authorization, \\
vaccine oppression,
vaccine satan,
covid vaccine bill gates, \\
covid vaccine side effect,
covid vaccine adverse events \\
     \hline
    \end{tabular}%
\end{adjustbox}
\caption{List of the keywords for data collection.} 
\label{tab:keywords}
\end{table}

\subsection{Data Annotation Task}\label{app:data_annotation_task}

\begin{figure*}[htbp]
  \centering  
  \includegraphics[width=  1 \textwidth]{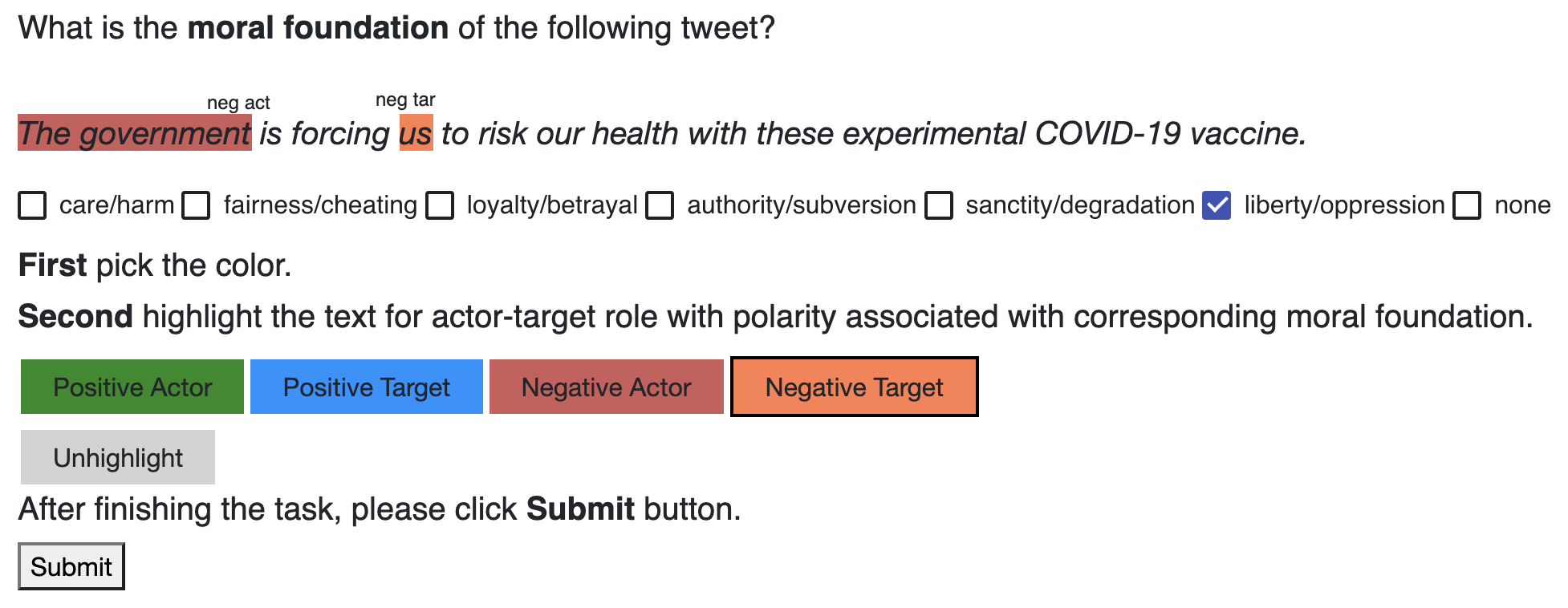}
    \caption{Annotation task interface.}
    \label{fig:ti}
\end{figure*}

The steps for annotating tweets using our graphical interface are (See Figure \ref{fig:ti}).
\begin{enumerate}
\item \textbf{Select} the moral foundation of the text using the checkbox $\text{\rlap{$\checkmark$}}\square$. You can see the definition of each moral foundation by hovering your mouse over them. If the tweet does not make any moral judgement, \textbf{check} $\text{\rlap{$\checkmark$}}\square$ "none". In this case, you don't have to highlight actor-target polarity.
 \item After selecting any moral foundation other than "none", text highlighting for actor-target role with polarity will be visible below. If you select a moral foundation other than "none", you can highlight the actor-target polarity.
\item \textbf{Choose} the color-coded label Positive Actor/Positive Target/Negative Actor/Negative Target to highlight the text with the color of the selected label. You can see the definition of actor-target-polarity role by hovering your mouse over them.
\item \textbf{Highlight} words, phrases, or sections of the text for the actor-target role with polarity of the corresponding moral foundation.
\item If you made any mistakes in highlighting, select the  "\textbf{Unhighlight}" button to unhighlight the previously highlighted text.
\item Finally, click the "\textbf{Submit}" button to submit the task.
\end{enumerate}

\begin{figure*}[htbp]
  \centering  
  \includegraphics[width=  1 \textwidth]{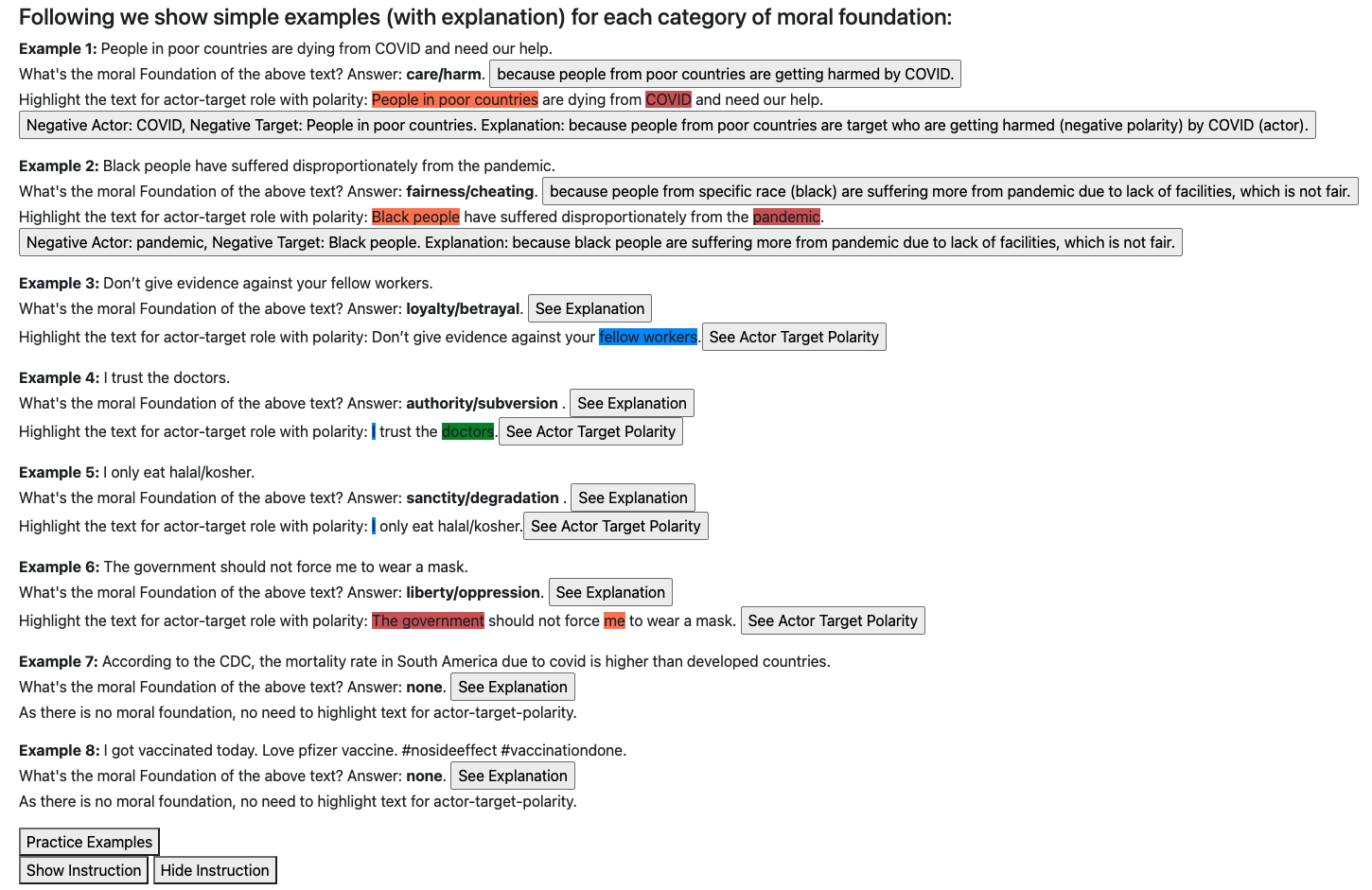}
    \caption{Examples provided to the annotators.}
    \label{fig:te}
\end{figure*}

We provided eight examples (Figure \ref{fig:te}) covering six moral principles and non-moral cases to make our annotation task more understandable. Annotators could see the explanations for choosing a moral foundation and an. actor-target polarity by clicking the  "\textbf{See Explanation}" button. 

\begin{figure*}[htbp]
  \centering  
  \includegraphics[width=  1 \textwidth]{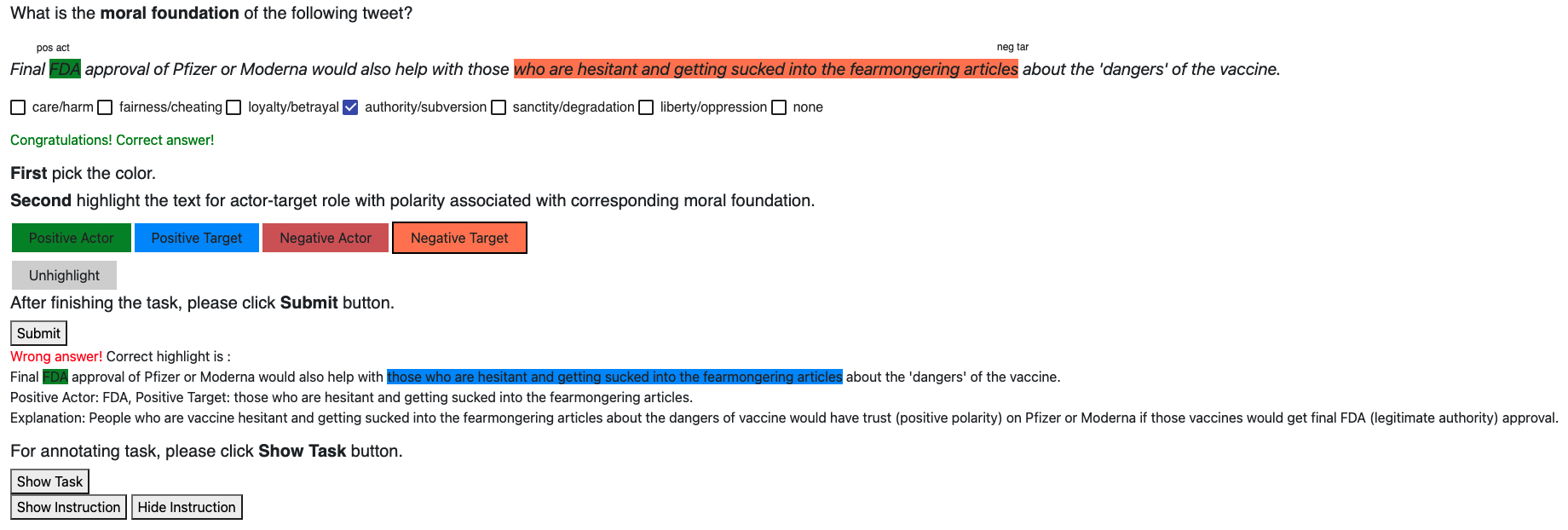}
    \caption{One of the two practice examples provided to the annotators before starting the real task.}
    \label{fig:tp}
\end{figure*}

Annotators had to complete two practice examples before starting the real task. If they made any mistake, our practice session provided them the correct result with an explanation. Figure \ref{fig:tp} shows the interface for one of the two practice examples.

\subsection{Out-of-Domain Datasets}\label{app:ood}

For moral foundation prediction, we use the dataset proposed by \citet{johnson-goldwasser-2018-classification}, consisting of 2K tweets by US congress members annotated for the five core moral foundations. We also use the Moral Foundation Twitter Corpus \cite{hoover2020mftc}, consisting of 35k tweets annotated for moral foundations. The topics across these two datasets span political issues (e.g. gun control, immigration) and events (e.g. Hurricane Sandy, Baltimore protests). Given that neither of these two datasets contain examples for the \textit{liberty/oppression} moral foundation, we curate a small lexicon by looking for synonyms and antonyms of the words \textit{liberty} and \textit{oppression}. Then, we use this lexicon to annotate the congresstweets dataset~\footnote{https://github.com/alexlitel/congresstweets}. We annotate a tweet as \textit{liberty/oppression} if it contains at least four keywords, which results in around 2K tweets. The derived lexicon for liberty/oppression can be seen in Table~\ref{tab:liberty}

To learn to predict roles, we use the subset of \citet{johnson-goldwasser-2018-classification} dataset annotated for roles by \citet{roy-etal-2021-identifying}, which contains roughly 3K tweet-entity-role triplets. For polarity, we combine the \citet{roy-etal-2021-identifying} dataset with the MPQA 3.0 entity sentiment dataset~\cite{deng-wiebe-2015-mpqa}, which contains about 1.6K entity-sentiment pairs. 

\begin{table}[H]
\begin{adjustbox}{width=\columnwidth, center}
\begin{tabular}{|l|}
\hline
liberty, independence, freedom, autonomy, sovereignty \\
self-government, self-rule, self-determination, home-rule \\
civil liberties, civil rights, human rights, autarky,\\ free-rein, latitude, option, choice, volition, democracy, \\
oppression, persecution, abuse, maltreatment, ill treatment,\\
dictator, dictatorship, autocracy, tyranny, despotism, \\
repression, suppression, subjugation, enslavement,\\
exploitation, dependence, constraint, control, totalitarianism \\
\hline
    \end{tabular}%
\end{adjustbox}
\caption{Liberty/Oppression Lexicon.} 
\label{tab:liberty}
\end{table}

For stance, we annotate our dataset of 85K unlabeled covid tweets using a set of prominent antivax and provax hashatgs. For the antivax case, we rely on the hashtags proposed by \citet{Muric2021}. For the provax case, we manually annotate hashtags that have a clear provax message, and that are used in at least 50 tweets in our unlabeled dataset. The full set of hashtags used can be found in Tables \ref{tab:pro_hashtags}  and \ref{tab:anti_hashtags}.

\begin{table}[t]
\resizebox{\columnwidth}{!}{%
\begin{tabular}{|l|}
\hline
FullyVaccinated, GetTheVax, GetVaccinatedASAP,\\ VaccineReady, VaxUpIL, TeamVaccine, GetTheJab,\\ VaccinesSaveLives, RollUpYourSleeve, DontMissYourVaccine,\\ letsgetvaccinated, TakeTheVaccine, takethevaccine,\\ COVIDIDIOTS, SafeVaccines, ThisIsOurShotCA,\\ LetsGetVaccinated, getthevaccine, GetVaccinated\\ PandemicOfTheUnvaccinated, VaccineStrategy, igottheshot, \\ vaccinationdone, ThisIsOurShot, VaccinateNiagara,\\ TwoDoseSummer, OurVaccineOurPride, IGotMyShot,\\ FreeVaccineForAll, VaccineEquity, COVIDIOTS, GetTheVaccine,\\
GetVaxxed, VaccineJustice, getthejab, VaccineForAll,\\ covidiot, gettheshot, RollUpYourSleevesMN, GoVAXMaryland,\\ WorldImmunizationWeek, VaccinesWork, getvaccinated,\\ GetVaccinatedNow, VaxUp, PlanYourVaccine,\\ VaccinateEveryIndian, TakeYourShot, Vaccines4All,\\ VaccinnateWithConfidence, firstdose, YesToCOVID19Vaccine,\\ NYCVaccineForAll, Vaccine4All, getvaxxed, VaccinEquity,\\ 
\hline
\end{tabular}}
\caption{ProVax Hashtags}\label{tab:pro_hashtags}
\end{table}

\begin{table}[t]
\resizebox{\columnwidth}{!}{%
\begin{tabular}{|l|}
\hline
abolishbigpharma, noforcedflushots, NoForcedVaccines,\\ ArrestBillGates, notomandatoryvaccines,\\ betweenmeandmydoctor, NoVaccine, bigpharmafia,\\ NoVaccineForMe, bigpharmakills, novaccinemandates,\\ BillGatesBioTerrorist, parentalrights, billgatesevil,\\ parentsoverpharma, BillGatesIsEvil, saynotovaccines,\\ billgatesisnotadoctor, stopmandatoryvaccination,\\ billgatesvaccine, cdcfraud, cdctruth, v4vglobaldemo, cdcwhistleblower\\  
vaccinationchoice, covidvaccineispoison, VaccineAgenda\\ 
depopulation, vaccinedamage, DoctorsSpeakUp, vaccinefailure,\\ educateb4uvax, vaccinefraud, exposebillgates, vaccineharm,\\ forcedvaccines, vaccineinjuries, Fuckvaccines, vaccineinjury \\ idonotconsent, VaccinesAreNotTheAnswer, informedconsent,\\ vaccinesarepoison, learntherisk, vaccinescause,\\ medicalfreedom, vaccineskill, medicalfreedomofchoice,\\ momsofunvaccinatedchildren, mybodymychoice \\
\hline
\end{tabular}}
\caption{AntiVax Hashtags} 
\label{tab:anti_hashtags}
\end{table}

\end{document}